\documentclass{article}

     \PassOptionsToPackage{numbers, compress}{natbib}
\usepackage[preprint]{neurips_2020}

\usepackage[utf8]{inputenc} % allow utf-8 input
\usepackage[T1]{fontenc}    % use 8-bit T1 fonts
\usepackage{hyperref}       % hyperlinks
\usepackage{amsmath}
\usepackage{mathtools}
\usepackage{multirow}
\usepackage{url}            % simple URL typesetting
\usepackage{booktabs}       % professional-quality tables
\usepackage{amsfonts}       % blackboard math symbols
\usepackage{nicefrac}       % compact symbols for 1/2, etc.
\usepackage{microtype}      % microtypography
\usepackage{amsthm}
\newtheorem{thm}{Theorem}

\title{Revisiting Oversmoothing in Deep GCNs}

 \author{%
   Chaoqi Yang, ~Ruijie Wang, ~Shuochao Yao, ~Shengzhong Liu, ~Tarek Abdelzaher\\
   University of Illinois, Urbana-Champaign, IL 61801, USA \\
   \texttt{\{chaoqiy2,ruijiew2,syao9,sl29,zaher\}@illinois.edu}
 }

\begin{document}
\maketitle

\begin{abstract}
	\textit{Oversmoothing} has been assumed to be the major cause of performance drop in deep graph convolutional networks (GCNs). 
%	However, the evidence is usually derived from simple graph convolution (SGC), a linear variant of GCNs. 
In this paper, we propose a new view that \textit{deep GCNs can actually learn to anti-oversmooth during training}. This work interprets a standard GCN architecture as  layerwise integration of a Multi-layer Perceptron (MLP) and graph regularization. We analyze and conclude that before training, the final representation of a deep GCN does over-smooth, however, it learns anti-oversmoothing during training. Based on the conclusion, the paper further designs a cheap but effective trick to improve GCN training.
	We verify our conclusions and evaluate the trick on three citation networks and further provide insights on neighborhood aggregation in GCNs.
% 	. The experiments show that (i) overfitting leads to the performance drop in deep GCN;
% 	,and oversmoothing does not exist even model goes to very deep (100 layers); 
% 	(ii) \textit{mean-subtraction} speeds up the model convergence as well as retains the same expressive power; (iii) the weight of neighbor averaging ($1$ is the common setting) does not significantly affect the model performance once it is above the threshold ($>0.5$).
\end{abstract}

\section{Introduction}

%- general, examples. 
%- good architecture GCN, GAT, ChebshevNet, GraphSage, 
%- theory side: discuss experssive power, 
%- oversmoothing problem, oversmoothing xxx.
%
%- xxx works on oversmoothing problem, their name of methods, xxx
%- from SGC, three main reasons.
%- oversmoothing do not exist. give figures.
%
%- we view it minmin-optimization, do not exist
%- we discuss mean-subtraction and discuss why it work
%- learing rate and weight of neighbor information xxx

Graph neural networks (GNNs) are widely used in modeling real-world graphs, like protein networks \cite{ying2018hierarchical}, social networks \cite{velivckovic2018deep}, and co-author networks \cite{kipf2016semi}.
One could also construct similarity graphs by linking data points that are close in the feature space even when there is no explicit graph structure. There have been several successful GNN architectures: ChebyshevNet \cite{defferrard2016convolutional}, GCN \cite{kipf2016semi}, 
% MPNN \cite{gilmer2017neural}, 
SGC \cite{wu2019simplifying}, GAT \cite{velivckovic2017graph}, GraphSAGE \cite{hamilton2017inductive} and other subsequent
variants tailored for practical applications \cite{gilmer2017neural,liu2019hyperbolic,klicpera2018predict}. 

Recently, researchers started to explore the fundamentals of GCNs \cite{kipf2016semi}, such as expressive power \cite{xu2018powerful,oono2019graph,loukas2019graph,dehmamy2019understanding}, and analyze their capacity and limitations. One of the frequently mentioned problem during GCN training is \emph{oversmoothing}~\cite{li2019deepgcns}. In deep graph convolution based architectures, \emph{over-smoothing} means that after multi-layer graph convolution, the effect of Laplacian smoothing makes node representations more and more similar, which eventually become indistinguishable. This issue was first mentioned in \cite{li2018deeper} and has been widely discussed since then, such as in JKNet~\cite{xu2018representation}, DenseGCN~\cite{li2019deepgcns}, DropEdge~\cite{rong2019dropedge}, and PairNorm~\cite{zhao2019pairnorm}. 

In this work, we propose a new understanding that \textit{though deep GCNs does lead to oversmoothing with initial parameters (before training), it can learn anti-oversmoothing during training.} This paper starts from the perspective of graph-based regularization model (two loss functions $\mathcal{L}_0 + \gamma \mathcal{L}_{reg}$ as supervision) \cite{belkin2003laplacian,yang2016revisiting}, where $\mathcal{L}_0$ is the empirical loss and $\mathcal{L}_{reg}$ is a graph regularizer, which encodes smoothness over the connected node pairs. 

We reformulate the MLP-based graph regularization model by two steps (and minimizing two loss functions respectively in each step), which gives the GCN model. STEP1 encodes the graph regularizer $\mathcal{L}_{reg}$ implicitly into the layerwise propagation of MLP, resulting in the GCN architecture (before renormalization);  STEP2 conducts standard back-propagation algorithm on $\mathcal{L}_{0}$, under the new architecture. Therefore, GCN could be expressed conceptually as a two-step minimization:
\begin{equation}
\underbrace{\mbox{STEP1 (layerwise):} \min_{\{X^{(l)}\}} ~~ \mathcal{L}_{reg}\left(\{X^{(l)}\}\mid \{W^{(l)}\}\right)}_{why~GCN~architecture?} ~~~~~~\mbox{and}~~~~~~\underbrace{\mbox{STEP2:} \min_{\{W^{(l)}\}}
	\mathcal{L}_0\left(\{W^{(l)}\}\right)}_{train~the~architecture.}. \notag
\end{equation}
% \begin{equation}\small
% \min_{\{W^{(l)}\}} \min_{\{X^{(l)}\}} ~~\mathcal{L}_0 \left(\{W^{(l)}\}\right) + \gamma \mathcal{L}_{reg}\left(\{X^{(l)}\}\mid \{W^{(l)}\}\right) ~~~~\mbox{and} ~~~~\min_{W} \min_X ~~\mathcal{L}_0 \left(W\right) + \gamma \mathcal{L}_{reg}\left(X\right) \notag
% \end{equation}
When viewed as a graph regularization problem, GCN actually encodes $\mathcal{L}_{reg}$ in the forward propagation rule (architecture) and train the parameters under the supervision of $\mathcal{L}_{0}$. 

\begin{figure}[t]
	\centering
	\includegraphics[width=5.5in]{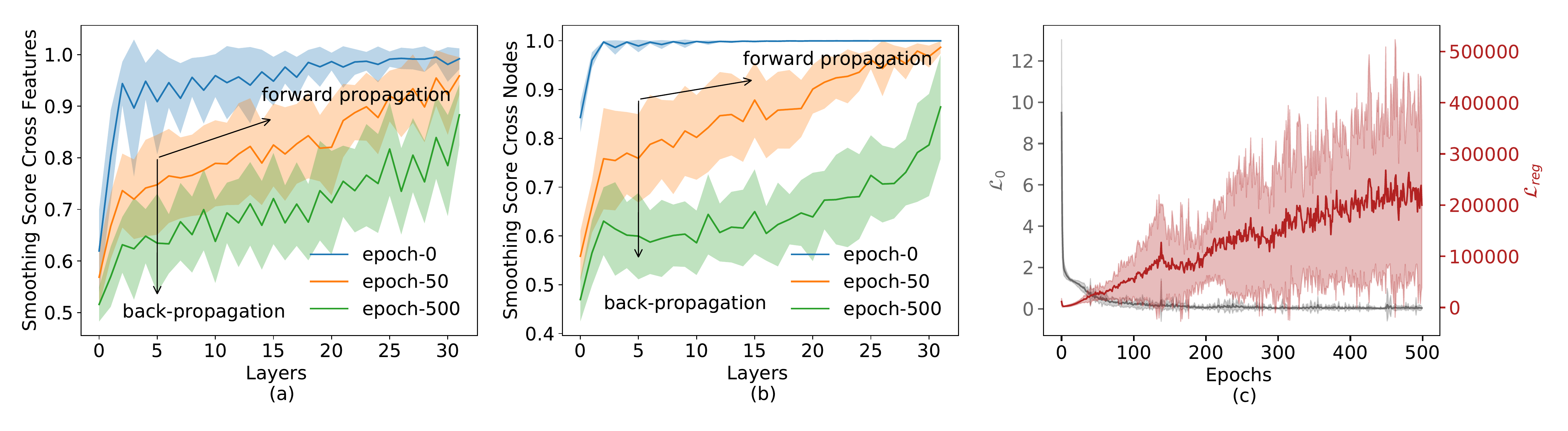}
	\vspace{-6mm}
	\caption{Demonstration of Anti-oversmoothing on \textit{Karate} Dataset. The \textit{Karate}  graph has 34 vertices of 4 classes (the same labeling strategy as \cite{kipf2016semi,perozzi2014deepwalk}) and 78 edges. Each class has two labeled samples. We apply 32-layer GCN with 16 hidden units on loss $\mathcal{L}_0$ (for 500 epochs). For the output of each layer, we compute the feature-wise smoothing score in Fig.~\ref{fig:karate}.(a) and node-wise smoothing score
	in Fig.~\ref{fig:karate}.(b). They are calculated by vector cosine similarity (details are given in Appendix~\ref{sec:AppendixF}). Fig.~\ref{fig:karate}.(c) shows two loss functions $\mathcal{L}_{0}$ and $\mathcal{L}_{reg}$ for each training epoch. Within one epoch, we observe that nodes/features are becoming more and more indistinguishable through forward propagation. For any layer, we observe that the smoothness  does disappear gradually during back-propagation/training. }
	\label{fig:karate}
	\vspace{-0.6mm}
\end{figure}

From this reformulation, we can clearly know that before training (after STEP1), deep GCNs do suffer from oversmoothing. Because the effect of deep GCN architecture will naturally minimize $\mathcal{L}_{reg}$, which gradually makes all the node representations proportional to the largest eigenvector of the {Laplacian} (we show it in Sec.~\ref{sec:gcnAsOptimization}). However, during STEP2/training, GCNs will learn to prevent oversmoothing because (i) the oversmoothing situation is conditioned on $\{W^{(l)}\}$; (ii) the explicit goal of STEP2/training is to find optimal $\{W^{(l)}\}$, so as to minimize empirical loss $\mathcal{L}_0$ and (iii) as long as the oversmoothing exists, feature representations will be indistinguishable, so upon minimizing $\mathcal{L}_0$, the model \emph{must} learn to make features separable, which naturally means anti-oversmoothing (see the demo in Figure~\ref{fig:karate}). 

Based on the reformulation, we further propose a \textit{mean-subtraction} trick. We show that applying mean-subtraction layerwise is equivalent to approximate the Fiedler vector (the second smallest eigenvector of the Laplacian), which set an initial graph partition and speeds up the  training.

%To prevent gradient vanishing/exploding, in this paper, we add skip connections \cite{he2016deep} to all deep architectures by default. An illustration of deep GCNs (with $\mathcal{L}_0$), deep SGC (with $\mathcal{L}_0$) and MLP (with $\mathcal{L}_0 + \gamma \mathcal{L}_{reg}$) is shown in Fig.~\ref{fig:GCNvsSGC}.
%
%\begin{figure}[t]
%	\centering
%	\includegraphics[width=5.5in]{fig/GCNvsSGC.pdf}
%	\caption{Comparison of Deep GCN, Deep SGC and DNN on \textit{Cora}}
%	\label{fig:GCNvsSGC}
%\end{figure}

In the experiment, we empirically verify that it is not oversmoothing that leads to a performance drop in training deep GCNs. Instead, we conjecture that \textit{overfitting} might be the major contributing factor based on experimental evidence.  The experiments also demonstrate the efficacy of the proposed mean-subtraction trick and provide more insights on neighborhood aggregation in GCNs.

%In the experiment, we empirically show that deep GCN does not present a typical oversmoothing pattern. Instead, we conjecture that \textit{overfitting} might be the major reason for performance drop based on the experimental evidence. Further experiments also demonstrate the efficacy of the proposed mean-subtraction trick and provide more insights on neighborhood aggregation in GCNs.

% We also explore  conduct As is mentioned above, the training of deep GCNs is a learning process of anti-oversmoothing, which is extremely slow in practice and sometimes may not converge. Based on the formulation, we further propose a \textit{mean-subtraction} trick to accelerate the training of deep GCNs. 
% % Compared to traditional BatchNorm \cite{ioffe2015batch} and GCN-related PairNorm \cite{zhao2019pairnorm}, our mean-subtraction retains the expressive power of deep GCNs and provides a speed-up as well. 
% Extensive experiments verify our theories and provide more insights about deep GCNs.

%\input{relatedworks}

\vspace{-0.4mm}
\section{Background of Graph Transductive Learning}
\vspace{-2mm}
Graph representation learning aims at embedding the nodes into low-dimensional vectors, while simultaneously preserving both \textit{graph topology structure} and \textit{node feature information}. 
%Afterwards, one then apply
%subsequent downstream learning algorithms for tasks, like node classification. 
%Transductive learning on graphs is the most significant one, where the inference is not only based on isolated features but also through node-to-node connections. 
Given a graph $G=(V,E)$, let $V=\{v_1, v_2, \cdots, v_n\}$ be the set of nodes,  and let $Y$ be a set of $m$ possible classes. Assume that each node $v_j$ is associated with a class label $y_j \in Y$. A graph could be represented by an \emph{adjacency matrix} $A$ with $A_{ij}=1$ when two nodes are connected $(v_i,v_j)\in E$. The \emph{degree matrix} $D=diag(d_1, d_2, \dots, d_n)$ is diagonal where $d_i=\sum_j A_{ij}$. Let
$X=\{x_1, x_2,\dots, x_n\}$ denote the feature vectors for each node. Given a labelled set $T\subset V$, the goal of transductive learning on a graph is to transductively predict labels for the remaining unknown nodes $V\setminus T$. A well-studied solution category is to include graph regularizers \cite{belkin2003laplacian,tenenbaum2000global,zhu2002learning,weston2012deep} into the classification algorithm. Graph convolution based
models~\cite{kipf2016semi,defferrard2016convolutional,velivckovic2017graph,hamilton2017inductive} are also powerful learning approaches in this space.

\subsection{Graph-based Regularization} \label{sec:reg}
\vspace{-2mm}
There is a rather general class of embedding algorithms that include graph regularizers. They could
be described as: finding a mapping $f(\cdot)$, i.e., MLP, by minimizing the following two-fold loss:
% \begin{equation}
% \mathcal{L} = \mathcal{L}_0 + \gamma \mathcal{L}_{reg} = \sum_{v_i \in T}\phi(f(x_i), y_i) + \gamma \sum_{A_{ij}\neq 0}\varphi (f(x_i), f(x_j), A_{ij}),
% \end{equation}
\vspace{0.6mm}
\begin{equation} \label{eq:graph-regularizer}
\mathcal{L} = \mathcal{L}_0\left(f(X)\right) + \gamma \mathcal{L}_{reg}\left(f(X)\right),
\end{equation}
\vspace{0.6mm}
where $f(X)=[f(x_i)]_{i=1}^n$ is the low-dimensional representation of nodes, and $\gamma$ denotes the weight. The first term is the empirical risk on the labelled set $T$. The second term is a graph regularizer over the connected pairs, so as to make sure that a trivial solution is not reached. 
% Hyperparameter $\gamma$ controls how much the algorithm will prefer $\mathcal{L}_{reg}$. $\phi(\cdot, \cdot)$ is a loss function between the prediction and target, while $\varphi(\cdot, \cdot, \cdot)$ quantifies the unsmoothness between a connected pair of nodes. 

The measurements on graphs are usually invariant to node permutations. A canonical way is to use Dirichlet energy \cite{belkin2002laplacian} for the graph-base regularization,
\begin{equation} \label{eq:trace}
\mathcal{L}_{reg} = \frac12 \sum_{i,j} A_{ij}\left\Vert\frac{f(x_i)}{\sqrt{d_i}}-\frac{f(x_j)}{\sqrt{d_j}}\right\Vert^2 = \frac{1}{2} ~\mbox{Tr}\left(f(X)^\top {\Delta}f(X)\right),
\end{equation}
where $\Delta= I - D^{-\frac12}AD^{-\frac12}$ is the \emph{normalized Laplacian operator}, which induces a semi-norm on $f(\cdot)$, penalizing the changes between adjacent vertices. Same normalized formulation could be found in \cite{chen2019deep,ando2007learning,smola2003kernels,shaham2018spectralnet,belkin2003laplacian}, and some related literature also use the unnormalized version \cite{zhu2002learning,von2007tutorial}.

\subsection{Graph Convolutional Network} \label{sec:gcn}
GCNs are derived from graph signal processing \cite{sandryhaila2013discrete,chen2015discrete,duvenaud2015convolutional}. On the spectral domain, the operator ${\Delta}$ is a real-valued symmetric semidefinite matrix and the graph convolution is parameterized by a learnable filter $g_\theta$ on  its eigenvalue matrix. Kipf et al. \cite{kipf2016semi} made
assumptions of the largest eigenvalue (i.e., $\lambda_{max}=2$) and simplified it with two-order Chebyshev expansion,
\begin{equation}
g_\theta \star x \approx \theta_0 x + \theta_1 ({\Delta} - I) x \approx \theta (I + D^{-\frac12}AD^{-\frac12})x.
\end{equation}

% GCN is derived from graph signal processing \cite{sandryhaila2013discrete,chen2015discrete,duvenaud2015convolutional}. The operator ${\Delta}$ is a real-valued symmetric semidefinite matrix and could be decomposed into
% ${\Delta}=U^\top\Lambda U$, where $U$ and $\Lambda$ are the matrix of eignvectors and eigenvalues.
% % ordered by eigenvalues and $\Lambda=diag(\lambda_1, \lambda_2,\dots,\lambda_n)$, $\lambda_1 > \cdots > \lambda_{i-1} > \lambda_n = 0$, is the eigenvalue matrix (spectrum). $U$ is in special orthogonal group $SO_n$ with $U^\top U =I$. 
% In spectral domain, the \emph{Fourier transform} of a signal $x$
% is defined by $\mathcal{F}(x)=U^\top x$ and its \emph{inverse transform} is $\mathcal{F}^{-1}(\hat{x}) = U\hat{x}$. The spectral graph convolution is parameterized by a learnable filter $g_\theta$ on the eigenvalues, i.e., $g_\theta(\Lambda)$, and could be approximated by truncated Chebyshev polynomials $T_k(x)$ up to $K_{th}$ order \cite{hammond2011wavelets},
% \begin{equation}
% g_\theta \star x = Ug_\theta(\Lambda) U^\top x \approx \sum_{k=0}^K \theta_k T_k({\Delta})x.
% \end{equation}
% This is the standard formulation of ChebyshevNet \cite{defferrard2016convolutional}. Kipf et al. \cite{kipf2016semi} made further
% assumptions of the largest eigenvalue (i.e., $\lambda_{max}=2$) and simplified it with two-order expansion,
% \begin{equation}
% g_\theta \star x \approx \theta_0 x + \theta_1 ({\Delta} - I) x \approx \theta (I + D^{-\frac12}AD^{-\frac12})x.
% \end{equation}

A multi-layer graph convolutional network (GCN) is formulated as the following layerwise propagation rule ($\sigma(\cdot)$ is an activation function, e.g., ReLU):
\begin{equation}
X^{(l+1)} =\sigma\left( \tilde{D}^{-\frac12}(I+A)\tilde{D}^{-\frac12}X^{(l)} W^{(l)}\right) \label{eq:rule}
\end{equation}
where $\tilde{D}^{-\frac12}(I+A)\tilde{D}^{-\frac12}\leftarrow I + D^{-\frac12}AD^{-\frac12}$ is the renormalization trick, $X^{(l)}$ and $W^{(l)}$ are the layerwise feature and parameter matrices, respectively.

\section{GCN as Layerwise  Integration of Graph Regularizer and MLP} \label{sec:gcnAsOptimization}
These two broad graph representation algorithms are closely related. In this section, we reformulate GCN  (in Sec.~\ref{sec:gcn}) from the MLP-based graph regularization algorithm (in Sec.~\ref{sec:reg}). %In this section, we re-interpret a GCN as a two-step optimization problem on standard MLP, where STEP1 is to minimize $\mathcal{L}_{reg}$ by viewing $\{W^{(l)}\}$ as constants and $\{X^{(l)}\}$ as parameters while STEP2 minimizes $\mathcal{L}_{0}$ by updating $\{W^{(l)}\}$. 
Essentially, we combine the MLP architecture and the gradient descent rule of minimizing $\mathcal{L}_{reg}$. We show that the resulting architecture is identical to the GCN before re-normalization.
Let us first discuss a gradient descent algorithm to minimize $\mathcal{L}_{reg}$.

% In \textit{Task1}, $X$ is treated as parameters to minimize $\mathcal{L}_{reg}$, while in \textit{Task2}, $X$ is fixed and the $\mathcal{L}_0$ will be minimized over layer-wise parameters $\{W^{(l)}\}$.

% \subsection{Layer-wise Propagation of MLP}
% Given the node set $V$, features $X=\{x_1, x_2,\dots, x_n\}$ and a labelled set $T\subset V$, a label mapping, $f_{\{W^{(l)}\}}: X \mapsto Y$, is usually a deep neural network, which could be tailored according to the practical applications. For example, $f(\cdot)$ could be a convolutional neural network (CNN) for image recognition or a recurrent neural network (RNN) for language processing. In this scenario, we first consider a simple multi-layer perceptron (MLP).
% % and define the cross-entropy loss over the labelled set $T$. Through back-propagation, layer-wise parameters $W^{(l)}$ will be updated to minimize $\mathcal{L}_0$.
% The forward propagation rule of an MLP is given by,
% \begin{equation} \label{eq:mlp}
% X^{(l+1)} = \sigma(X^{(l)}W^{(l)}),
% \end{equation}
% where $W^{(l)}$ and $X^{(l)}$ are layer-wise parameters and inputs. 

\subsection{Gradient Descent for Minimizing $\mathcal{L}_{reg}$} \label{sec:graidentDescent}

Given the Laplacian operator $\Delta \in\mathbb{R}^{n\times n}$, we consider to minimize the graph regularizer $\mathcal{L}_{reg}=\frac12 ~\mbox{Tr}(X^\top \Delta X)$ on feature domain $X \in \mathbb{R}^{n\times d}$, where $d$ is the input dimension. To prevent the trivial solution $X=\textbf{0}\in \mathbb{R}^{n\times d}$, we consider the energy constraint on $X$, i.e., $\|X\|^2_F = c_1\in \mathbb{R}^+$. The trace optimization problem is:
\begin{equation}
\min ~~\frac12 ~\mbox{Tr}(X^\top \Delta X),~subject~to~const.~\|X\|^2_F,
\end{equation}
where $\|X\|_F^2$ denotes the Forbenius-norm  of $X$. To solve this, We equivalently transform the optimization problem into the \emph{Reyleigh Quotient} form $R(X)$, which is,
\begin{equation}
\min ~~R(X)=\frac{\frac12 ~\mbox{Tr}(X^\top \Delta X)}{\|X\|^2_F}=\frac{\frac12 ~\mbox{Tr}(X^\top \Delta X)}{\mbox{Tr}(X^\top X)},~subject~to~const.~\|X\|^2_F.
\end{equation}
It is obvious that $R(X)$ is scaling invariant on $X$, i.e., $\forall ~c_2 \neq 0 \in \mathbb{R}$, $R(X)=R(c_2\cdot X)$. 

\paragraph{One-step Improvement.} Given an initial guess, $X$, one-step of trace optimization aims at finding a better guess $X_{better}$, which satisfies $R(X_{better})\leq R(X)$ and $\|X_{better}\|_F^2=c_1$. Our strategy is first viewing the problem as unconstrained optimization on $R(X)$ and update the guess $X$ to the intermediate value $X_{mid}$, such that $R(X_{mid}) \leq R(X)$, by gradient descent. Then we rescale $X_{mid}$ to reach the improved guess $X_{better}$, which meets the norm constraint. 

Given the initial guess $X$, we apply gradient descent with learning rate $\eta=\tiny\frac{\mbox{Tr}(X^\top X)}{ 2-\frac{\mbox{Tr}(X^\top\Delta X)}{\mbox{Tr}(X^\top X)}}$ and reach an intermediate solution $X_{mid}$ in the unconstrained space:
\begin{align}
\nabla_{X} &= \frac{\partial R(X)}{\partial X} = \frac12\frac{\partial \frac{\mbox{Tr}(X^\top \Delta X)}{\mbox{Tr}(X^\top X)}}{\partial X} = \frac{\left(\Delta - I\frac{\mbox{Tr}(X^\top \Delta X)}{\mbox{Tr}(X^\top X)}\right)X}{\mbox{Tr}(X^\top X)}, \\
%X' &= \left. X^{(l)} -\eta \nabla_{X}\right|_{X=X^{(l)}} = \frac{(2-\Delta)X^{(l)}}{2-\frac{\mbox{Tr}(X^{(l)\top}\Delta X^{(l)})}{\mbox{Tr}(X^{(l)\top} X^{(l)})}} = \frac{(I + D^{-\frac12}AD^{-\frac12})}{2-\frac{\mbox{Tr}(X^{(l)\top}\Delta X^{(l)})}{\mbox{Tr}(X^{(l)\top} X^{(l)})}}X^{(l)},
X_{mid} &= X -\eta \nabla_{X} = \frac{(2-\Delta)X}{2-\frac{\mbox{Tr}(X^{\top}\Delta X)}{\mbox{Tr}(X^{\top} X)}} = \frac{(I + D^{-\frac12}AD^{-\frac12})}{2-\frac{\mbox{Tr}(X^{\top}\Delta X)}{\mbox{Tr}(X^{\top} X)}}X.
\end{align}
Immediately, we get $R(X_{mid}) \leq R(X)$ (proofs in Appendix~\ref{sec:appendixB}). Then, we rescale $X_{mid}$ by a constant $c_3\in \mathbb{R}^+$, i.e., $X_{better}=c_3\cdot X_{mid}$, so as to meet the norm constraint, i.e., $\|X_{better}\|_F^2=c_1$. 

\paragraph{Discussion of the Form.}In sum, we reach a better guess $X_{better}$, which satisfies $R(X_{better}) =R(X_{mid})\leq R(X)$ and follows this form,
\begin{equation} \label{eq:betterGuess}
 X_{better}= c_3\cdot X_{mid} \propto (I + D^{-\frac12}AD^{-\frac12})X.
\end{equation}

Note that, the eigenvectors of $(I + D^{-\frac12}AD^{-\frac12})$ and $\Delta$ are the same. The operator  $(I +D^{-\frac12}AD^{-\frac12})$ in Equation~\eqref{eq:betterGuess} is similar to an one-step \emph{Laplacian} smoothing, which explains why it will give a better guess in terms of minimizing $\mathcal{L}_{reg}$. However, it  causes the issue of \textit{oversmoothing} after sufficient number of layers. We will discuss in depth in Sec.~\ref{sec:oversmoothing}. It is also interesting that when combining with MLP architecture, the only magic in Equation~\eqref{eq:betterGuess} is the operator $(I +D^{-\frac12}AD^{-\frac12})$, since the scalar will be absorbed into the layerwise parameter matrix of MLP. Let us discuss this below.

\subsection{Layerwise Propagation and Optimization}
We introduce the solution from Sec.~\ref{sec:graidentDescent} into the layerwise propagation of MLP. Given the node set $V$, features $X=\{x_1, x_2,\dots, x_n\}$ and a labelled set $T\subset V$, a label mapping, $f_{\{W^{(l)}\}}: X \mapsto Y$, is usually a deep neural network, which could be tailored according to the practical applications. In this scenario, we consider a simple multi-layer perceptron (MLP).
% and define the cross-entropy loss over the labelled set $T$. Through back-propagation, layer-wise parameters $W^{(l)}$ will be updated to minimize $\mathcal{L}_0$.
The forward propagation rule of an standard MLP is given by,
\begin{equation} \label{eq:mlp}
X^{(l+1)} = \sigma(X^{(l)}W^{(l)}), ~~l = 1,\dots,L
\end{equation}
where $X^{(0)}$ is the feature matrix, $W^{(l)}$ and $X^{(l)}$ are layerwise parameters and inputs. 

\paragraph{STEP1: minimizing $\mathcal{L}_{reg}$ in Forward Propagation.} To ensure that the output of MLP could lead to a smaller $\mathcal{L}_{reg}$, an intuitive way is to apply the gradient descent step between the layerwise propagation, so that the smoothing effect will be accumulated layer-by-layer towards the final representation. Let us consider the output of the $(l-1)$-th layer, i.e., $X^{(l)}$. We know from Sec.~\ref{sec:graidentDescent} that through one-step gradient descent, it transforms into:
\begin{align} \label{eq:constant}
X^{(l)}_{better} \propto (I + D^{-\frac12}AD^{-\frac12})X^{(l)}.
%X^{(l+1)} &= \sigma(X'W^{(l)}) = \sigma((I + D^{-\frac12}AD^{-\frac12})X^{(l)}W'^{(l)}), \label{eq:combine2}
\end{align}
We plug this new value into Eqn.~\eqref{eq:mlp} and immediately reach the same convolutional propagation rule, $f^{(l+1)}:X^{(l)} \mapsto  X^{(l+1)}$, as Kipf et al. \cite{kipf2016semi} (before applying the renormalization trick),
\begin{equation}
X^{(l+1)} = \sigma(X^{(l)}_{better}W^{(l)}) = \sigma\left((I + D^{-\frac12}AD^{-\frac12})X^{(l)}W^{(l)}_{new}\right),
\end{equation}
where the constant scalar in Eqn.~\eqref{eq:constant} is absorbed into parameter matrix $W^{(l)}$, resulting in $W^{(l)}_{new}$ (we still use the notation $W^{(l)}$ below if there is no ambiguity). 
Therefore, a GCN forward propagation is essentially applying STEP1 layerwise in the forward propagation of an MLP, which is a composition of mappings $f = f^{(L)} \circ \cdots \circ f^{(1)}$ on initial feature $X^{(0)}$. In essence, the GCN structure implicitly contains the goal of minimizing graph regularizer.

\paragraph{STEP2: minimizing $\mathcal{L}_{0}$ in Back Propagation.} From the above, we have transformed the graph regularization $\mathcal{L}_{reg}$ as a layerwise convolution operator. Then the empirical loss $\mathcal{L}_0$ will be the only supervision. In the model training process, the standard back-propagation algorithm is used on $\mathcal{L}_0$.

\subsection{GCN: combining STEP1 and STEP2} \label{sec:two-stepopt}
In sum, the GCN model can be interpreted from a graph regularization view. STEP1 encodes the graph regularizer implicitly into an MLP propagation, which explains "why GCN architecture?" In STEP2, under that architecture, the optimal $\{W^{(l)}\}$ is learned and a low-dimension $f(X)$ is reached with respect to $\mathcal{L}_0$ explicitly and $\mathcal{L}_{reg}$ implicitly, after standard loss back-propagation.  When viewed as a graph regularization problem, GCN could be expressed conceptually as a two-step optimization,
\begin{equation} \label{eq:min-min}
\underbrace{\mbox{STEP1 (layerwise):} \min_{\{X^{(l)}\}} ~~ \mathcal{L}_{reg}\left(\{X^{(l)}\}\mid \{W^{(l)}\}\right)}_{why~GCN~architecture?} ~~~~~~\mbox{and}~~~~~~\underbrace{\mbox{STEP2:} \min_{\{W^{(l)}\}}
\mathcal{L}_0\left(\{W^{(l)}\}\right)}_{train~the~architecture.}.
\end{equation}

% given a feature vector, left production by a convolution operator is equivalent to minimizing $\mathcal{L}_{reg}$, and right production by $W^{(l)}$ together with the non-linear activation is equivalent to one-layer perceptron. After standard back-propagation algorithm, the optimal $\{W^{(l)}\}$ is learned and a low-dimension $X^{(L)}$ is reached with respect to both $\mathcal{L}_0$ and $\mathcal{L}_{reg}$.  We express it as a min-min optimization,
% \begin{equation} \label{eq:min-min}
% \min_{\{W^{(l)}\}} \min_X ~~\mathcal{L}_0 \left(\{W^{(l)}\}\right) + \gamma \mathcal{L}_{reg}\left(X\mid \{W^{(l)}\}\right).
% \end{equation}
In this section, the learning rate $\eta=\tiny\frac{\mbox{Tr}(X^\top X)}{ 2-\frac{\mbox{Tr}(X^\top\Delta X)}{\mbox{Tr}(X^\top X)}}$ is specially chosen, and it satisfies $\eta \in(0, \infty)$ since $X^\top X$ is semi-definite and $\frac{\mbox{Tr}(X^\top\Delta X)}{\mbox{Tr}(X^\top X)}$ is smaller than the largest eigenvalue of $\Delta$, which is smaller than $2$. 
In the experiment section, we reveal that $\eta$ is related to the weight of neighborhood aggregation. We further test different $\eta$ and provide more insights on how to set the aggregation weights in Sec.~\ref{sec:lr} experiments. In the following sections, we use $A_{sym}$ to denote
the re-normalized convolutional operator $\tilde{D}^{-\frac12}(I+A)\tilde{D}^{-\frac12}$ and use $A_{rw}$ for the random walk form $\tilde{D}^{-1}(I+A)$.

\section{Analysis and Improvement} \label{sec:oversmoothing}
\vspace{-2mm}
The recent successes in applying GNNs are largely limited to \emph{shallow} architectures (e.g., 2-4 layers).
Model performance decreases when adding more intermediate layers. Summarized in
\cite{zhao2019pairnorm}, there are three possible contributing factors: (i) overfitting due to increasing number
of parameters; (ii) gradient vanishing/exploding; (iii) oversmoothing due to Laplacian smoothing. The first two points are common in all deep architectures. The issue of  oversmoothing
is therefore our focus in the section. We analyze the behavior of deep GCNs  in terms of the training process and conclude that \textit{the training process of GCN starts from the oversmoothing situation, and deep GCNs can learn to anti-oversmooth} (in the experiment, we show that \textit{overfitting} might be  the major factor for performance drop).
%Our conclusion is that \textit{deep GCNs can learn anti-oversmoothing by nature, but overfitting is the major cause of performance drop}.
 Based on the analysis, we further propose
 a cheap but effective trick to speed up deep GCNs training.

 \subsection{Conditional Over-smoothing Before Training}  
 \vspace{-1mm}
 \emph{Oversmoothing} means that node representations become more and more similar and finally go indistinguishable after multi-layer graph convolution. Previous literature \cite{li2018deeper,xu2018representation,rong2019dropedge,li2019deepgcns,nt2019revisiting,zhao2019pairnorm} already discussed that due to the Laplacian smoothing effect, deep GCN architectures lead to oversmoothing. The primary reason is summarized in Theorem~\ref{thm:1} (see proofs in Appendix~\ref{sec:AppendixD}). \cite{oono2019graph} further provides a similar result when considering the ReLU activation function during the analysis, under the assumption that the  singular values of each parameter matrix are bounded by $1$.
 
 \begin{thm} \label{thm:1}
	Given any random signal $x\in\mathbb{R}^n$ and a symmetric matrix $A\in\mathbb{R}^{n\times n}$, the following property $\displaystyle{\lim_{k \to \infty}} A^kx \propto u_1 $ holds almost everywhere on $x$, where $A$ has non-negative eigenvalues and $u_1$ is the eigenvector associated with the largest eigenvalue of $A$.
\end{thm}

Specifically, for two widely used convolution operators $A_{sym}$ and $A_{rw}$, they have the same dominant eigenvalue $\lambda_1 = 1$ with eigenvectors $\tilde{D}^{\frac12} \textbf{1}$ and $\textbf{1}$, respectively. Before training, it is intuitive that if the depth $L$ goes to infinity, then the  output channel of GCN will become proportional to $\tilde{D}^{\frac12}\textbf{1}$ or $\textbf{1}$. 

Luckily, this situation is conditioned on the parameter matrices. During training, the learned parameters can substantially reverse the smoothing effect through the non-linear functions (in Appendix~\ref{sec:analysisSGC}, we analyze SGC \cite{wu2019simplifying} similarly, which is a linear version of GCN. We prove that without the activation function, the oversmoothing of SGC will be independent of the parameters and thus SGC  cannot learn to anti-oversmooth). During training, GCN will learn to address the oversmoothing issue gradually.

 \subsection{Anti-oversmoothing During Training} \label{sec:anti-oversmoothing}
 \vspace{-1mm}
 We show in Sec.~\ref{sec:gcnAsOptimization} that GCN could be expressed as a two-step optimization, where the STEP1 minimizes $\mathcal{L}_{reg}$ conditioned on $\{W^{(l)}\}$, and the STEP2 finds optimal $\{W^{(l)}\}$ and minimizes $\mathcal{L}_0$. In this section, we shall analyze why STEP2 will learn anti-oversmoothing naturally.
 
 \paragraph{Analysis.} Before training the GCN architecture, we know from Theorem~\ref{thm:1} that the output representation of each feature channel will be proportional to the largest eigenvector of the convolution operator. In that case, the feature representations are indistinguishable and lead to a small graph regularization loss $\mathcal{L}_{reg}$ but a large supervised loss $\mathcal{L}_0$ (like Figure~\ref{fig:karate}.(c)).  The training process of GCN is to re-balance the trade-off between $\mathcal{L}_{reg}$ and $\mathcal{L}_{0}$ from the initial oversmoothing situation.
 
$\mathcal{L}_{reg}$ encodes the smoothness over the connected node pairs, which favors the solution, where the connected nodes share similar representations ("similar" means the scale of each feature channel is approximately proportional to the square root of its degree, refer to Equation~\eqref{eq:trace}, and oversmoothing is an extreme case). $\mathcal{L}_{0}$ calculates the error based on labels. If the labels are well-aligned with node degree information, then $\mathcal{L}_{reg}$ and $\mathcal{L}_{0}$ will aim to learn similar representations and the feature matrices will not change a lot during training. However, in real practice,  the labels usually contain other semantic information, so that the goals of two loss functions, i.e., $\mathcal{L}_{reg}$ and $\mathcal{L}_{0}$, are not always aligned. Since the explicit supervision of STEP2/training is $\mathcal{L}_0$, the training process is actually a step-by-step shift from the oversmoothing region (where node representations are proportional to the square root of the degrees) towards the optimal region (where node representations are partially similar, separable and well-aligned with the labeling semantics), which naturally means anti-oversmoothing. 
 
% \paragraph{Non-linearity.} In fact, the ability of learning anti-oversmoothing is ensured by the non-linear functions in GCN architecture. The change of the parameter set $\{W^{(l)}\}$ can substantially reverse the Laplcian smoothing effect through the activation functions. Let us consider the linear version, Simplified Graph Convolution (SGC) \cite{wu2019simplifying}, where the activation function is removed (we primarily consider the ReLU function). Suppose $A$ is a convolution operator, the $L$-layer SGC is of the form, $f(X) = A^{L}XW$, where $W = \prod_{i=0}^{L-1}W^{(L-1)}$ is an aggregated parameter matrix. We similarly reformulate SGC as a two-step optimization,
% \begin{equation}\label{eqn:min-min-SGC}
% \mbox{STEP1:} ~ \min_{X} ~ \mathcal{L}_{reg}(X) ~~~~~~\mbox{and}~~~~~~\mbox{STEP2:} ~\min_{W}~
% \mathcal{L}_0 (W).
% \end{equation}
% It is intuitive that without the non-linear functions, STEP1 and STEP2 are independent in SGC model.
% In the experiment (Sec.~\ref{sec:exp1}), we compare SGC with GCN models. We show that without non-linear functions, SGC is hard to get anti-oversmoothing effect and its performance drops dramatically when model layers go up. However, with the dependency between STEP1 and STEP2, GCNs can learn anti-oversmoothing and give good performance when model goes deep. 

 %%%%%%%%%%%%%%%%%%%%%%%%%%%%%%%%%%%%%%%%%%
  \subsection{Improve Deep GCN Training} \label{sec:meansubtraction}
   \vspace{-1mm}
  As revealed above, the learning of deep GCN begins with the oversmoothing situation, which makes the training slow.
 %   To improve deep GCNs, previous literature \cite{li2018deeper,xu2018representation,rong2019dropedge,li2019deepgcns,nt2019revisiting,zhao2019pairnorm} have started some insightful trials.
 %   However, their primary goal is to alleviate the oversmoothing ``problem''. 
   This issue has not been explored extensively in the literature \cite{chen2018fastgcn,chiang2019cluster}. In this work, we propose a cheap but effective trick to ensure a better beginning point and accelerate GCN training.
%    of deep GCNs, which theoretically magnifies the effect of Fiedler vector. 
%   PairNorm \cite{zhao2019pairnorm} also includes a \textit{mean-subtraction} step, however, their purpose is to simplify derivation.
  
  \vspace{-1.5mm}
  \paragraph{Motivation.}We propose \textit{mean-subtraction}, i.e., reducing the mean value from each feature channel of each hidden layer. Our motivation primarily stems from Theorem~\ref{thm:1}, where the \textit{Power Iteration} of convolution operator is to approximate the largest eigenvector, which causes an oversmoothing start. After applying  \textit{mean-subtraction}, the revised \textit{Power Iteration} will lead to the Fiedler vector (the second smallest eigenvector), which provides a coarse graph partition result and makes the training faster. PairNorm \cite{zhao2019pairnorm} also includes a \textit{mean-subtraction} step, however, the authors 
 did not state extensively in their paper. Our paper instead analyzes the mechanism.

 %  mean-subtraction are usually used together with standard deviation scaling. The effect is 
 %  to stabilize the learning process and reduce the number of training epochs.
\vspace{-1.5mm}
 \paragraph{Mean-subtraction.}We start with operator $A_{rw}$ and its largest eigenvector $u_1 = \textbf{1}\in \mathbb{R}^n$. For any  output feature channel $k$ of the $l$-th layer, i.e., $X_k^{(l)}\in\mathbb{R}^{n}$, the \textit{mean-subtraction} gives,
 \begin{equation} \label{eq:mean_subtraction}
 X^{(l)}_{k} \leftarrow X^{(l)}_k-\bar{X}^{(l)}_{k} = X^{(l)}_k - \frac{\textbf{1} \textbf{1}^\top X^{(l)}_k}{n} = X^{(l)}_k - \langle X^{(l)}_k, \bar{u_1}\rangle\cdot \bar{u_1}
 \end{equation}
 where $\bar{u_1}=\frac{u_1}{\|u_1\|}$. Eqn.~\eqref{eq:mean_subtraction} essentially reduces the components aligned with $\{u_1\}$-space. This is exactly one-step approximation of the Fiedler vector by \textit{Power Iteration}. Fiedler vector is widely used to partition a graph \cite{chung1997spectral} (demo on \textit{Karate} in Appendix~\ref{sec:AppendixF}) in spectral graph theory, and it seperates nodes initially. In essence, the vanilla GCN models train from the oversmoothing stage. With \textit{mean-subtraction} trick, the revised GCNs will consequently train on a coarse graph partition result, which is much faster. For the symmetric operator $A_{sys}$, the formulation will be adjusted by a factor, $\tilde{D}^{\frac12}$ (refer to derivation in  Appendix~\ref{sec:AppendixE}).  We show the power of \textit{mean-subtraction} in Sec.~\ref{sec:exp-mean-subtraction}.

%%%%%%%%%%%%%%%%%%%%%%%%%%%%%%%%%%%%%%%%%%

% Besides, practical implementation usually includes a biased term in the convolution layer (i.e., $b$ is the bias for feature projection $w^\top x + b$), and \textit{mean-subtraction} actually moves the mean signal into the biased term (i.e, $w^\top x+b\rightarrow w^\top(x-x_{mean}) + (b+w^\top x_{mean})$). Thus, the \textit{mean-subtraction} operation retains the representation power of GCNs.
% We demonstrate the efficiency in experiments.

\vspace{-1mm}
\section{Experiments}
\vspace{-2mm}
In this section, we present experimental evidence on \textit{Cora}, \textit{Citeseer}, \textit{Pubmed} to answer the following questions: (i) what is the real cause of performance drop in deep GCNs and why? (ii) How to improve (accelerate and stabilize) the training of a generic deep GCN model?  (iii) Does the learning rate $\eta$ (defined in Sec.~\ref{sec:gcnAsOptimization}) matter? How to choose the weights of neighborhood aggregation?

\vspace{-1mm}
\paragraph{Experiment Setup.} The experiments are basically on semi-supervised node classification tasks. We use \textit{ReLU} as the activation function. All the deep models (with more than 3 hidden layers) are implemented with skip-connection \cite{kipf2016semi,he2016deep}, since skip-connection (also called residual connection)
are necessary to prevent gradient exploding/vanishing in deep architectures, and we do not consider them as new models. We add the output of $l$-th layer to $(l+2)$-th layer after the \textit{ReLU} function. Three benchmark datasets (\textit{Cora}, \textit{Citeseer}, \textit{Pubmed}) are considered. We follow the same experimental settings from \cite{kipf2016semi} and show the basic statistics of datasets in 
Table.~\ref{tb:graphstat}. All the experiments are conducted with \textit{PyTorch 1.4.0} for 20 times and mainly finished in a Linux server with 64GB memory,  32 CPUs and two GTX-2080 GPUs. Details and additional experiments could be found in Appendix~\ref{sec:additionalexp}.

\begin{table}[htbp] \footnotesize
	\caption{Overview of Citation Network Statistics}
	\vspace{-1.5mm}
	\centering
	\begin{tabular}{ c|ccccc} 
			\toprule
			\textbf{Dataset}  & \textbf{\#Nodes}  & \textbf{\#Edges}  & \textbf{\#Features} & \textbf{\#Class} & \textbf{Label rate} \\
			\midrule
			Cora     & 2,708  & 5,429  & 1,433    & 7       & 0.052       \\
			Citeseer & 3,327  & 4,732  & 4,732    & 6       & 0.036      \\
			Pubmed   & 19,717 & 44,338 & 500      & 3       & 0.003   \\  
			\bottomrule
	\end{tabular}
	\label{tb:graphstat}
	\vspace{-2.5mm}
\end{table}

\subsection{Overfitting in Deep GCNs} \label{sec:exp1}
\vspace{-2mm}
The performance of GCNs is known to decrease with increasing number of layers, for which, a common explanation is \textit{oversmoothing} \cite{li2018deeper}. In Sec.~\ref{sec:oversmoothing}, we already analyze that deep GCNs can learn to anti-oversmooth. Empirically, this section further conjecture that \textit{overfitting might be the major reason for the drop of performance in deep GCNs}. We start with a comparison between the accuracy curve of GCN and SGC, and the latter one shows a typical oversmoothing pattern.
\vspace{-2mm}

\vspace{-1mm}
\paragraph{Oversmoothing or Not.} SGC \cite{wu2019simplifying} is a linear version of GCN without activation function. We show the accuracy curves of deep GCNs (with $\mathcal{L}_0$) and deep SGC (with $\mathcal{L}_0$) on \textit{Cora} and \textit{Pubmed} for various model depths in Fig.~\ref{fig:GCNvsSGC2}. An analysis of oversmoothing for SGC can be found in Appendix~\ref{sec:analysisSGC}. It is interesting that the accuracy of SGC decreases rapidly \cite{oono2019graph} with more graph convolutions either for training or test. This is a strong indicator of oversmoothing. The performance of the GCN model is not as good as SGC soon after 2 layers (because of overfitting possibly), but it stabilizes at a high accuracy even as the model goes very deep, which presents a non-oversmoothing pattern.

\begin{figure}[t]
	\centering
	\includegraphics[width=5.5in]{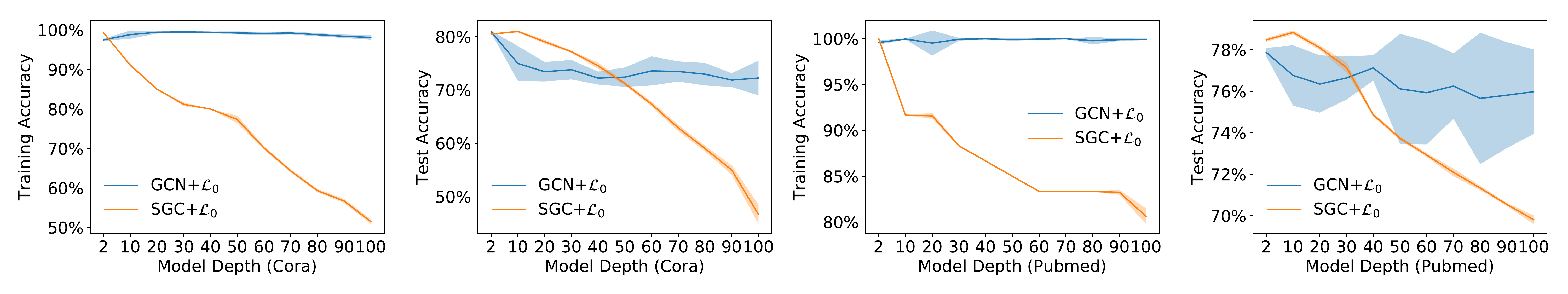}
	\vspace{-8mm}
	\caption{Comparison of Deep GCN and Deep SGC on \textit{Cora} and \textit{Pubmed}}
	\label{fig:GCNvsSGC2}
	\vspace{-3mm}
\end{figure}

\vspace{-2mm}
\paragraph{Loss Function vs Depth.} For a further investigation, we compute the training and test loss of the vanilla GCN models (with residual connection) on \textit{Cora, Citeseer} and \textit{Pubmed}. The loss curve of 2-, 3-, 5-, 10-, 50-layer GCNs with 1000 epochs on \textit{Cora} is reported in Figure~\ref{fig:loss} (reader could find similar \textit{Citeseer} and \textit{Pubmed} figures in Appendix~\ref{sec:additionalexp}). We notice that for shallow models (2- or 3-layer GCN), both the training and the test curve goes down with more epochs. However in deeper GCNs, the training curve almost hits the ground and the test curve first decreases and then increases gradually with more epochs (note that while the test loss increases, the test accuracy remains stable, reader could refer to Appendix~\ref{sec:additionalexp}). We therefore conclude that  \textit{overfitting} might be the major factor that leads to the performance drop in deep GCNs. Note that the test loss is almost horizontal for a 3-layer GCN, so we think 3 (or 4) layers might be a separation between overfitting or not, which is consistent with the common understandings that 2 or 3 layer-GCN works better than deep GCNs in most cases.

%The results
%of training and test accuracy are reported in Fig.~\ref{fig:threesets}. Form the figure, we notice that in the beginning (2-4 layers), these is a big rise in training accuracy (up to $\geq 99\%$) and simultaneously a big drop
%in training loss (to $\leq0.1$) and test accuracy consistently on three datasets. Then from 4 to 50 layers (we think 50 layers are deep enough), model performance is relatively stable (maybe drop a little bit because of numerical sensitivity or instability in deep architectures). We conclude that this evidence  is more consistent with \textit{overfitting}, which explains why shallow GCN works better than deep GCN in most cases.

\begin{figure}[t]
	\centering
	\includegraphics[width=5.5in]{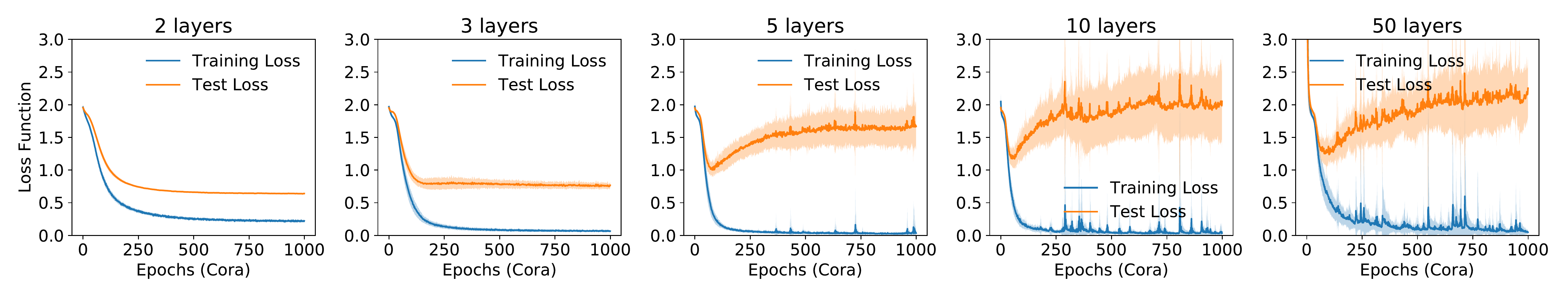}
	\vspace{-7mm}
	\caption{Training and Test Curve with 2-, 3-, 5-, 10-, 50-layer GCNs on \textit{Cora}}
	\label{fig:loss}
	\vspace{-2mm}
\end{figure}

%\paragraph{Deep GCNs Learn Anti-oversmoothing.}
%We recall that in Sec.~\ref{sec:anti-oversmoothing}, we show from an optimization perspective that the dependency of $\mathcal{L}_{reg}$ on $\{W^{(l)}\}$ allows the network to learn anti-oversmoothing.
%%Our formulation is compatible with the common understanding that SGC will suffer from oversmoothing, but our theory further shows that GCN could learn to anti-oversmoothing. 
%To verify our theory, we compare SGC and GCN on \textit{Cora} and \textit{Pubmed} with various depth. To make it clear, SGC is actually a linear model, the depth $L$ means the number of graph convolution $S^L$. Model performance in both training and test is shown in Fig.~\ref{fig:GCNvsSGC2}.
%
%It is interesting that the accuracy of SGC decreases rapidly \cite{oono2019graph} with more graph convolutions either for training or test. This is a strong indicator of oversmoothing, because node features converge to the same stationary due to the effect of STEP1 (specified in Theorem~\ref{thm:1}). The performance of the GCN model is not as good as SGC soon after 2 layers because of overfitting, but it stabilizes at a high accuracy even as the model goes very deep, which again verifies that GCNs naturally have the power of anti-oversmoothing.

%\begin{figure}[t]
%	\centering
%	\includegraphics[width=5.5in]{fig/GCNvsSGC2.pdf}
%	\caption{Comparison of Deep GCN and Deep SGC on \textit{Cora} and \textit{Pubmed}}
%	\label{fig:GCNvsSGC2}
%\end{figure}
\vspace{-2mm}
\subsection{Mean-subtraction for GCNs} \label{sec:exp-mean-subtraction}
\begin{figure}[t]
	\centering
	\includegraphics[width=5.3in]{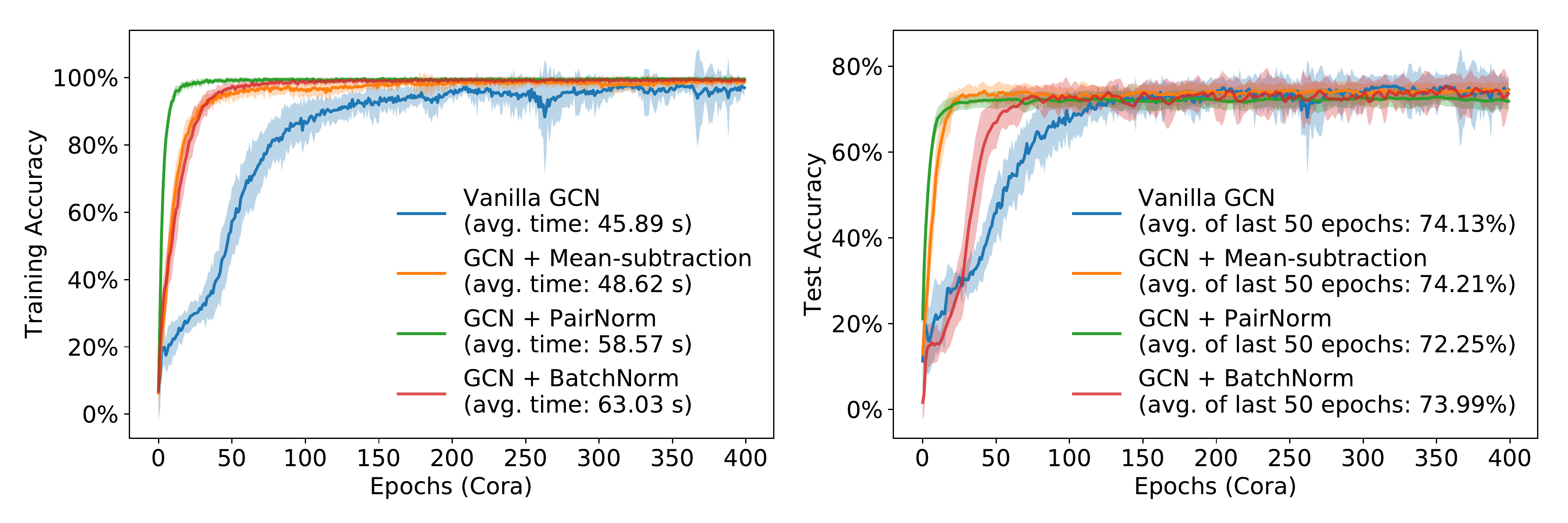}
	\vspace{-3mm}
	\caption{Comparison of Different Tricks in Training Deep GCNs on \textit{Cora}.}
	\label{fig:mean}
	\vspace{-2mm}
\end{figure}

\vspace{-2mm}
In this section, we evaluate the efficacy of the \textit{mean-subtraction} trick and compare it with vanilla GCNs~\cite{kipf2016semi}, \textit{PairNorm}~\cite{zhao2019pairnorm} and the commonly used \textit{BatchNorm}~\cite{ioffe2015batch}. The four models have the same configurations, such as the number of layers (64), epochs (400), learning rate (0.01), and hidden units (16). They differ in how to transform the layerwise feature matrices. \textit{Mean-subtraction} is to subtract
the mean feature value before each convolution layer, and \textit{PairNorm} will add a re-scaling step on the top. These two tricks
do not include additional parameters. \textit{BatchNorm} includes more parameters for each layer, which learns the mean and variance of feature representation. The experiment is conducted for all three datasets. In this section, we plot for \textit{Cora} (in Fig.~\ref{fig:mean}) due to space limitation. Readers could find similar \textit{Citeseer} and \textit{Pubmed} curves in Appendix~\ref{sec:additionalexp}.

\vspace{-2mm}
\paragraph{Analysis.} We report the \textit{Cora} result in Fig.~\ref{fig:mean}, with average time consumption for 400 epochs and average accuracy of last 50 epochs. After all, \textit{mean-subtraction}, \textit{PairNorm} and \textit{BatchNorm} all help to improve the training process: (i) fit training data well (see high training accuracy); (ii) give fast convergence. Compared to
\textit{BatchNorm},  \textit{mean-subtraction} provides a robust/stable training curve (small variance) with less executed time. Compared to \textit{PairNorm}, our \textit{mean-subtraction} outputs a higher accuracy efficiently in test data. We think that additional re-scaling step in \textit{PairNorm} might cause severe overfitting problem. In sum, \textit{mean-subtraction} not only speeds up the model convergence but also retains the same expressive power. It is an ideal trick for training deep GCNs.

\subsection{Weight of Neighborhood Aggregation in GCNs} \label{sec:lr}
% \begin{figure}[t]
% 	\centering
%	\includegraphics[width=0.7\textwidth]{fig/neighbor.pdf}
%	\caption{Different Averaging Weight for 2-layer and 32-layer GCN in \textit{Cora}}
%	\label{fig:neighbor}
%\end{figure}
In Sec.~\ref{sec:gcnAsOptimization}, we choose the learning rate $\eta=\tiny\frac{\mbox{Tr}(X^\top X)}{ 2-\frac{\mbox{Tr}(X^\top\Delta X)}{\mbox{Tr}(X^\top X)}}$. However, a different learning rate does lead to different weights $w(\eta)$ of neighbor information aggregation (we show that $w(\eta)$ is a monotonically increasing function in Appendix~\ref{sec:AppendixC}). There are also some efforts on trying different ways to aggregate neighbor information \cite{velivckovic2017graph,hamilton2017inductive,rong2019dropedge,chen2018fastgcn}. In this section, we consider the form "$I +w(\eta) A$" with $w(\eta)\in [0,\infty)$ and exploit a group of convolution operators by their normalized version. GCN with normalized $I +w(\eta) A$ is named as $\eta$-GCN. We evaluate this operator group on \textit{Cora} and list the experimental results in Table.~\ref{tb:neighbor}

%\begin{table}[htbp] \scriptsize
%	\centering
%	\begin{tabular}{ c|c|cccccccccccc} 
%		\toprule
%		\multicolumn{2}{c|}{\textbf{Accuracy (\%)}}  & \textbf{$w(\eta)$=0}  & \textbf{0.1}  & \textbf{0.2} & \textbf{0.5} & \textbf{0.8} & \textbf{1.0}  & \textbf{2}  & \textbf{5} & \textbf{10} & \textbf{20} & \textbf{50} & \textbf{100}\\
%		\midrule
%		\multirow{2}{*}{2-layer}    & training  & 92.84&96.22&96.62&96.56&95.87&95.38&94.91&93.51&93.47&92.84&92.78&92.88     \\
%		& test  & 51.86&71.36&75.62&78.93&79.62&79.87&79.04&78.50&78.19&77.97&77.71&77.79    \\
%		\midrule 
%		\multirow{2}{*}{32-layer}   & training & 94.78&99.52&99.70&99.35&99.13&99.31&98.46&98.37&98.35&98.38&98.26&98.47  \\  
%		& test & 40.81&73.08&75.01&73.40&74.48&74.38&76.34&75.50&73.70&76.36&76.08&75.01     \\
%		\bottomrule
%	\end{tabular}
%	\caption{Performance vs Neighbor Averaging Weight (2-layer and 32-layer)}
%	\label{tb:neighbor}
%\end{table}

\begin{table}[htbp] \scriptsize
	\caption{Performance vs Neighborhood Aggregation Weight (2-layer and 32-layer) on \textit{Cora}}
	\vspace{-2mm}
	\centering
	\begin{tabular}{ c|c|ccccccccccc} 
		\toprule
		\multicolumn{2}{c|}{\textbf{Accuracy (\%)}}  & \textbf{$w(\eta)$=0}  & \textbf{0.1}  & \textbf{0.2} & \textbf{0.5} &  \textbf{1.0}  & \textbf{2}  & \textbf{5} & \textbf{10} & \textbf{20} & \textbf{50} & \textbf{100}\\
		\midrule
		\multirow{2}{*}{2-layer}    & training  & 92.66&95.67&96.32&96.05&95.33&94.54&93.44&93.30&92.82&92.86&92.98   \\
		& test  & 50.75&74.99&78.11&80.38&81.23&80.90&79.82&80.01&80.50&79.77&79.10    \\
		\midrule 
		\multirow{2}{*}{32-layer}   & training & 95.02&99.49&99.58&99.35&98.69&98.10&98.84&98.83&98.81&98.76&98.83  \\  
		& test & 39.93 & 72.53 & 73.59 & 73.65 & 74.03 & 75.11 & 74.96 & 75.08 & 75.49 & 74.64 & 74.74    \\
		\bottomrule
	\end{tabular}
	\label{tb:neighbor}
	\vspace{-4mm}
\end{table}

\paragraph{Analysis.}We conclude that when $w(\eta)$ is small (i.e., $\eta$ is small), which means the gradient of $\mathcal{L}_{reg}$ does not contribute much to the end effect,  $\eta$-GCN is more of a DNN. As $w(\eta)$ increases, a significant increase in model performance is initially observed. When $w(\eta)$ exceeds some threshold, the accuracy saturates, remaining high (or maybe decreases slightly) even as we increase $w(\eta)$ substantially. We conclude that for the widely used shallow GCNs, the common choice of weight $w(\eta)=1$, which means a learning rate, $\eta=\tiny\frac{\mbox{Tr}(X^\top X)}{ 2-\frac{\mbox{Tr}(X^\top\Delta X)}{\mbox{Tr}(X^\top X)}}$, is large enough to include the $\mathcal{L}_{reg}$ gradient descent effect and small enough to avoid the drop in accuracy. For a deeper GCN model, larger weight ($>1.0$) is preferable. To find the best weight of neighbor averaging, further inspection is needed in future work.

\section{Conclusion}
\vspace{-2mm}
In this work, we reformulate the graph convolutional networks (GCNs) from MLP-based graph regularization algorithm. Based on that, we analyze the training process of deep GCNs and provide a new understanding: \textit{deep GCNs have the power to learn anti-oversmoothing by nature, and overfitting might be the major reason for the performance drop when model goes deep}. We further propose a cheap but effective \textit{mean-subtraction} trick to accelerate the training of deep GCNs. Extensive experiments are presented to verify our theories and provide more practical insights.

%\section*{Broader Impact}
%This paper provides deep understandings and improve the training of deep graph convolutional networks (GCNs). The paper has the following potential positive impact in the society: our work could encourage/improve the usage of deep learning models on various graph related applications. In the meantime, while this paper is primarily theoretical, it might cause possible negative consequences when the assumptions are not satisfied or the graph structures are not constructed correctly from the real-world applications.

%\section*{Broader Impact}
%
%Authors are required to include a statement of the broader impact of their work, including its ethical aspects and future societal consequences. 
%Authors should discuss both positive and negative outcomes, if any. For instance, authors should discuss a) 
%who may benefit from this research, b) who may be put at disadvantage from this research, c) what are the consequences of failure of the system, and d) whether the task/method leverages
%biases in the data. If authors believe this is not applicable to them, authors can simply state this.
%
%Use unnumbered first level headings for this section, which should go at the end of the paper. {\bf Note that this section does not count towards the eight pages of content that are allowed.}

\bibliographystyle{unsrt}
\bibliography{main.bib}

\newpage
%%%
\appendix 

\section{$\mathcal{L}_{reg}$ and Spectral Clustering}
\paragraph{Graph Regularizer $\mathcal{L}_{reg}$.} $\mathcal{L}_{reg}$ is commonly formulated by Dirichlet energy, $\mathcal{L}_{reg} = \frac{1}{2} ~\mbox{Tr}\left(H^\top {\Delta}H\right)$, where $f(\cdot)$ is a mapping from the input feature $X$ to 
low-dimensional representation $H=f(X)$. To minimize $\mathcal{L}_{reg}$, this paper adds constraint on the magnitude of $H$, i.e., $\|H\|_F^2=C\in \mathbb{R}$, which gives,
\begin{equation} \label{eq:appendixA}
\min_H ~\frac{1}{2} ~\mbox{Tr}\left(H^\top {\Delta}H\right), ~subject~to~const.~\|H\|^2_F.
\end{equation}

\paragraph{Spectral Clustering.} Given a graph with binary adjacency matrix $A$, a partition of node set $V$ into $k$ set could be written as $P_1, P_2, \dots, P_k$ in graph theory. For normalized spectral clustering, the $k$ indicator vectors is written as $h_i=(h^1_i,\dots, h^n_i)$, where
$h_i^j$ represents the affiliation of node $j$ in class set $P_i$ and $vol(P_i)=\sum_{v_j\in P_i}d_j$ is the volume.
\begin{equation}
h^j_i = \left\{\begin{array}{rr}
\frac{1}{\sqrt{vol(P_i)}}, & v_j~\in P_i \\
0, & otherwise
\end{array}\right.
\end{equation}
The $H=[h_i^j]_{i=1..k,j=1..n}$ is a matrix containing these $k$ indicator vectors as columns. For each row of
$H$, there is only one non-empty entry, implying $h_i^\top h_j=0,~\forall i\neq j$. Let us revisit the \textit{Normalized Cut} of a graph for a partition $P_1, P_2, \dots, P_k$.
\begin{align}
\mbox{Ncut}(P_1, P_2, \dots, P_k) &= \sum_{i=1}^k\frac{cut(P_i, \bar{P_i})}{vol(P_i)} \notag\\
&= \frac12 \sum_{i=1}^k \frac{\sum_{v_j\in P_i, v_t \notin P_i}A_{jt} + \sum_{v_j\notin P_i, v_t \in P_i}A_{jt}}{\sqrt{vol(P_i)}\sqrt{vol(P_i)}} \notag\\
&= \frac12 \sum_{i=1}^k\left(\sum_{v_j\in P_i, v_t \notin P_i}\left(\frac{1}{\sqrt{vol(P_i)}}\right)^2+ \sum_{v_j\notin P_i, v_t \in P_i}\left(\frac{1}{\sqrt{vol(P_i)}}\right)^2 \right) \notag\\
& = \frac12 \sum_{i=1}^k \sum_{j,t}(h_i^j - h_i^t)^2 = \frac12 \sum_{i=1}^kh_i^\top L h_i = \frac 12 \mbox{Tr} (H^\top L H).
\end{align}
Also, $H$ satisfies $H^\top DH=I$. When the discreteness condition is relaxed and  $H$ is substitute by $H=D^{-\frac12}U$, the normalized graph cut problem (normalized spectral clustering) is relaxed into,
\begin{equation} \label{eq:appendixA2}
\min_U ~\frac{1}{2} ~\mbox{Tr}\left(U^\top {\Delta}U\right), ~~subject~to~~~U^\top U =I.
\end{equation}
This is a standard trace minimization problem which is solved by the matrix the eigen matrix of $\Delta$. Compared to Eqn.~\eqref{eq:appendixA}, Eqn.~\eqref{eq:appendixA2} has a stronger constraints, which outputs the optimal solution irrelevant to the inputs (feature matrix $X$). However, Eqn.~\eqref{eq:appendixA} only add constraints on the magnitude of $H$, which balances the trade-off and will give a solution induced by both the eigen matrix of $\Delta$ and the original feature $X$.

\section{Reyleigh Quotient $R(X)$} \label{sec:appendixB}

\paragraph{Reyleigh Quotient.} The Reyleigh Quotient of a vector $x\in \mathbb{R}^m$ is the scalar,
\begin{equation}
R(x) = \frac{x^\top \Delta x}{x^\top x},
\end{equation}
which is invariant to the scaling of $x$. For example, $\forall ~c_1 \neq 0 \in \mathbb{R}$, we have $R(x)=R(c_1\cdot x)$. When we view $R(x)$ as a function on $m$-dim variable $x$, it has stationary points $x_i$, where $x_i$ is the eigenvector of $\Delta$. Let us assume $\Delta x_i = \lambda_i x_i$, then the stationary value at point $x_i$ will  be exactly the eigenvalue $\lambda_i$,
\begin{equation}
R(x_i) = \frac{x_i^\top \Delta x_i}{x_i^\top x_i} = \frac{x_i^\top \lambda_i x_i}{x_i^\top x_i} = \lambda_i.
\end{equation}

When $x$ is not an eigenvector of $\Delta$, the partial derivatives of $R(x)$ with respect to the vector
coordinate $x_j$ is calculated as,
\begin{align}
\nabla_{x_j} R(x) &= \frac{\partial R(x)}{\partial x_j} = \frac{\frac{\partial}{\partial x_j}(x^\top \Delta x)}{x^\top x} - \frac{(x^\top \Delta x)\frac{\partial}{\partial x_j} (x^\top x)}{(x^\top x)^2} \notag\\
& = \frac{2(\Delta x)_j}{x^\top x} - \frac{(x^\top \Delta x)2x_j}{(x^\top x)^2} = \frac{2}{x^\top x} (\Delta x - R(x)x)_j
\end{align}
Thus, the derivative of $R(x)$ with respect to $x$ is collected as,
\begin{equation} \label{eq:AppexdixB}
\nabla R(x) = \frac{2}{x^\top x} (\Delta x - R(x)x).
\end{equation}

\paragraph{Minimizing R(x).} Suppose $\Delta= I - D^{-\frac12}AD^{-\frac12}$ is the normalized Laplacian matrix. Let us first consider to minimize $R(x)$ without any constraints.  Since  $\Delta$ is a symmetric real-valued matrix, it could be factorized by Singular Value Decomposition,
\begin{equation}
\Delta = U\Lambda U^\top = \sum_{i=1}^su_i \lambda_i u_i^\top
\end{equation}
where $s$ is the rank of $\Delta$ and $0=\lambda_1\leq \cdots \leq \lambda_s < 2$ are the eigen values. For any non-zero vector $x$, it is decomposed w.r.t. the eigen space of $\Delta$,
\begin{equation}
x =\epsilon + \sum_{i=1}^sc_i\cdot u_i
\end{equation}
where $\{c_i\}$ is the coordinates and $\epsilon$ is a component tangent to the eigen space spanned by $\{u_i\}$. Let us consider the component of $x$ within the eigen space and discuss $\epsilon$ later.
Therefore, the Reyleigh Quotient $R(x)$ can be calculated by,
\begin{equation}
R(x) =  \frac{x^\top \Delta x}{x^\top x} = \frac{(\sum_{i=1}^sc_i\cdot u_i^\top) (\sum_{i=1}^su_i \lambda_i u_i^\top)(\sum_{i=1}^sc_i\cdot u_i)}{( \sum_{i=1}^sc_i\cdot u_i^\top)(\sum_{i=1}^sc_i\cdot u_i)} = \frac{\sum_{i=1}^s c_i^2 \lambda_i}{\sum_{i=1}^s c_i^2}
\end{equation}
Recall the partial derivative of $R(x)$ w.r.t. $x$ in Eqn.~\eqref{eq:AppexdixB}. Think about to minimize $R(x)$ by gradient descent and always consider the learning rate (the same as what we used in the main text. The factor $\frac12$ is from that the $R(x)$ in appendix does not have the scalar $\frac12$),
\begin{equation}
\eta = \frac12 \mbox{Tr}(X^\top X) / (2-\frac{\mbox{Tr}(X^\top\Delta X)}{\mbox{Tr}(X^\top X)}) = \frac12 \cdot \frac{x^\top x}{2-\frac{x^\top \Delta x}{x^\top x}} = \frac12\cdot \frac{x^\top x}{2-R(x)}.
\end{equation}
The initial $x$ is regarded as an starting point, and the next point $x'$ is given by gradient descent,
\begin{equation}
x' = x - \eta \nabla R(x) = x - \frac12 \cdot \frac{x^\top x}{2-R(x)} \frac{2}{x^\top x} (\Delta x - R(x)x) = \frac{2I - \Delta}{2-R(x)} x.
\end{equation}
The new Reyleigh Quotient value is,
\begin{equation}
R(x') = \frac{x'^\top \Delta x'}{x'^\top x'} = \frac{{(\frac{2I - \Delta}{2-R(x)}x)}^\top \Delta {(\frac{2I - \Delta}{2-R(x)}x)}}{{(\frac{2I - \Delta}{2-R(x)}x)}^\top {(\frac{2I - \Delta}{2-R(x)}x)}} = \frac{x^\top (2I - \Delta)\Delta(2I - \Delta) x}{x^\top (2I - \Delta)(2I - \Delta)x}.
\end{equation}
The eigen properties of $2I-\Delta$ could be derived from $\Delta$, where they have the same eigenvector, and any eigenvalue $\lambda$ of $\Delta$ will adjust to be an eigenvalue $2-\lambda$ of $2I-\Delta$. Therefore, we do further derivation,
\begin{equation}
R(x') = \frac{\sum_{i=1}^s c_i^2(2-\lambda_i)^2 \lambda_i}{\sum_{i=1}^s c_i^2(2-\lambda_i)^2}.
\end{equation}
So far, to get the ideal effect, a final check is needed: whether the Reyleigh Quotient does decrease after the gradient descent.
\begin{align}
R(x') - R(x) &= \frac{\sum_{i=1}^s c_i^2(2-\lambda_i)^2 \lambda_i}{\sum_{i=1}^s c_i^2(2-\lambda_i)^2} -\frac{\sum_{i=1}^s c_i^2 \lambda_i}{\sum_{i=1}^s c_i^2} \notag\\
& = \frac{(\sum_{i=1}^s c_i^2)(\sum_{i=1}^s c_i^2(2-\lambda_i)^2 \lambda_i) - (\sum_{i=1}^s c_i^2(2-\lambda_i)^2)(\sum_{i=1}^s c_i^2 \lambda_i) }{(\sum_{i=1}^s c_i^2(2-\lambda_i)^2)(\sum_{i=1}^s c_i^2)} \notag\\
& = \frac{\sum_{i,j}c_i^2c_j^2(\lambda_i-\lambda_j)(\lambda_j-\lambda_i)(4-\lambda_i-\lambda_j)}{(\sum_{i=1}^s c_i^2(2-\lambda_i)^2)(\sum_{i=1}^s c_i^2)} \notag\\
& = - \frac{\sum_{i,j}c_i^2c_j^2(\lambda_i-\lambda_j)^2(4-\lambda_i-\lambda_j)}{(\sum_{i=1}^s c_i^2(2-\lambda_i)^2)(\sum_{i=1}^s c_i^2)} < 0
\end{align}

Also, we show the asymptotic property of $R(x)$ in gradient descent,
\begin{equation} \label{eq:AppendixB2}
\displaystyle{\lim_{t \to \infty}} R(x^{(t)}) = \displaystyle{\lim_{t \to \infty}} \frac{\sum_{i=1}^s c_i^2(2-\lambda_i)^{2t} \lambda_i}{\sum_{i=1}^s c_i^2(2-\lambda_i)^{2t}} = \frac{c_{2}^2\lambda_2}{c_{1}^2} \cdot \displaystyle{\lim_{t \to \infty}}  \left(\frac{2-\lambda_2}{2-\lambda_1}\right)^{2t}= 0^+
\end{equation}
where $x^{(t)}$ is the $t$-th new point given by gradient descent. So far, we finish the proof of well-definedness of gradient descent with the $\eta=\tiny\frac{x^\top x}{ 2-\frac{x^\top\Delta x}{x^\top x}}=\frac{\mbox{Tr}(X^\top X)}{ 2-\frac{\mbox{Tr}(X^\top\Delta X)}{\mbox{Tr}(X^\top X)}}$.

\paragraph{Remark 1.} In fact, as stated above, $R(x)$ is invariant to the scaling of $x$, so we could scale $x$ on its magnitude, i.e., making $\|x\|=c\in\mathbb{R}^+$ as a constraint during the gradient descent iteration, all the properties and results still hold. 

\paragraph{Remark 2.} In the main text, instead of using a vector $x$, we use a feature matrix $X$ and define our 
Reyleigh Quotient by $R(X) = \frac{\mbox{Tr}(X^\top \Delta X)}{\mbox{Tr}(X^\top X)}$. In fact, different feature channels of $X$ could be viewed as independent vector signal $x_i\in\mathbb{R}^m$ and for each channel, the same gradient descent analysis is applied. Therefore, we finish the detailed proof for our formulation in the main text, which is of the following form,
\begin{equation}
\min ~R(X),~subject~to~const.~\|X\|_F^2.
\end{equation}

%$\displaystyle{\lim_{k \to \infty}} A_{sym}^kx \propto \tilde{D}^{\frac12}\textbf{1}\in\mathbb{R}^n$ or $\displaystyle{\lim_{k \to \infty}} A_{rw}^kx \propto \textbf{1}\in\mathbb{R}^n$
%Mean-subtraction and Fiedler Vector

\section{Learning Rate $\eta$ and Neighbor Averaging Weight $w(\eta)$} \label{sec:AppendixC}

We show the relation of learning rate $\eta$ and neighbor averaging weight $w(\eta)$ in this section (to make the derivation consistent with the main text, $\nabla R(x)$ does not have factor $2$).
\begin{align}
X' &= X - \eta \nabla R(X) = X- \eta \frac{1}{\mbox{Tr}(X^\top X)} (\Delta X - R(X)X) \\
&= (\frac{R(X)-1}{\mbox{Tr}(X^\top X)}\cdot \eta + 1) (I + \frac{\eta}{(R(x) - 1)\cdot \eta + \mbox{Tr}(X^\top X)}\cdot D^{-\frac12}AD^{-\frac12})X
\end{align}
The first multiplier $(\frac{R(X)-1}{\mbox{Tr}(X^\top X)}\cdot \eta + 1)$ will be absorbed into the parameter matrices. Thus, we have,
\begin{equation}
w(\eta) = \frac{\eta }{(R(x) - 1)\cdot \eta + \mbox{Tr}(X^\top X)} = \frac{1}{R(x) - 1 + \frac{\mbox{Tr}(X^\top X)}{\eta}},
\end{equation}
According to the formulation, $w(\eta)$ is a monotonically increasing function on variable $\eta$ and is valid when $w(\eta)>0$. Therefore, when $R(x) \geq 1$, the domain of the function is $\eta \in [0, \infty)$ and when $R(x) < 1$ (we know from Eqn.~\eqref{eq:AppendixB2} that $R(x) \rightarrow 0^+$), the domain of the function is bounded, $\eta \in [0, \frac{\mbox{Tr}(X^\top X)}{1-R(x)})$. 

\paragraph{Remark 3.} Note that the choice in this paper, $\eta=\tiny\frac{\mbox{Tr}(X^\top X)} {2-\frac{\mbox{Tr}(X^\top\Delta X)}{\mbox{Tr}(X^\top X)}}$, always lies in the valid domain for $\forall x\in \mathbb{R}^m$. Also, in the valid domain,
with respect to the change of $\eta$, $w(\eta)$ can vary in the range $[0, \infty)$ monotonically.

\section{Proof of Theorem 1} \label{sec:AppendixD}

\begin{proof}
	 Given any non-zero signal $x\in\mathbb{R}^n$ and a symmetric matrix $A\in\mathbb{R}^{n\times n}$ (with non-negative eigenvalues), we factorize them in the eigenspace,
\begin{equation}
A = U\Lambda U^\top = \sum_{i=1}^su_i \lambda_i u_i^\top~~~\mbox{and}~~~x =\epsilon + \sum_{i=1}^sc_i\cdot u_i
\end{equation}
where $S$ is of rank $s\in\mathbb{N}^+$, $U=[u_i]_{i=1}^s$ and $\Lambda=diag(\lambda_1,\cdots, \lambda_s)$ are eigen matrices. $\{c_i\in\mathbb{R}\}_{i=1}^s$ are coordinates of $x$ in the eigenspace and $\epsilon\in\mathbb{R}$ is
a component tangent to the eigenspace. In a $k$-layer SGC, the effect of graph convolution is the same as applying Laplacian smoothing $k$ times,
\begin{equation}
A^kx = \left(\sum_{i=1}^su_i \lambda_i u_i^\top\right)^k\left(\epsilon + \sum_{i=1}^sc_i\cdot u_i\right) = \sum_{i=1}^sc_i\lambda_i^k u_i.
\end{equation}
Suppose $\lambda_1$ is the largest eigenvalue and $c_1\neq 0$. We go with infinite number of layers and then have $\displaystyle{\lim_{k \to \infty}} S^kx \propto u_1 $, which means the output is unrelated to the input features $x$.
\end{proof}

\section{Analysis of SGC} \label{sec:analysisSGC}
 SGC was proposed in \cite{wu2019simplifying}, with the hypothesis that the non-linear activation is not critical while the majority of the benefit arises from the local averaging $A$. The authors directly remove the activation function and proposed a linear ``$L$-layer'' model, where $\prod W^{(l)}$ has collapsed into a single $W$.  
\begin{equation} \label{eq:linear}
f(X) = A\left(A(\cdots)W^{(L-1)}\right)W^{(L)} = A^{L}XW.
\end{equation}

This model explicitly disentangles the dependence of STEP1 and STEP2. We similarly formulate the SGC model in the form of two-step optimization,
\begin{equation}\label{eqn:min-min-SGC}
\mbox{STEP1:} ~ \min_{X} ~ \mathcal{L}_{reg}(X) ~~~~~~\mbox{and}~~~~~~\mbox{STEP2:} ~\min_{W}~
\mathcal{L}_0 (W)
\end{equation}
From the two-step optimization form, SGC is essentially conducting gradient descent algorithm $L$ times in STEP1. In STEP2, SGC model will seek to minimize $L_{0}$ on the basis of the oversmoothed features. The independence between STEP1 and STEP2 accounts for the oversmoothing issue, which cannot be mitigated during training.

\section{Mean-subtraction} \label{sec:AppendixE}
\paragraph{Background.} In the main text, we start with one of the most popular convolution operator $A_{rw}$ and its largest eigenvector $u_1 = \textbf{1}\in \mathbb{R}^n$. Let us use a simplified notation $\bar{u_1}=\frac{u_1}{\|u_1\|}$. Given any non-zero $x\in\mathbb{R}^n$, the proposed mean-subtraction has the following form,
\begin{equation}
x_{new} \leftarrow x-x_{mean} = x - \frac{\textbf{1} \textbf{1}^\top x}{n} = x - \langle x, \bar{u_1}\rangle\cdot \bar{u_1}
\end{equation}
There are some facts from spectral graph theory that
\begin{itemize}
    \item $A_{sys}$ and $A_{rw}$ have the same eigenvalues, $1 = \lambda_1 \geq \cdots \lambda_s > 0$.
    \item If $u$ is an eigenvector of $A_{sys}$, i.e., $A_{sys}u = \lambda u$. Then $\tilde{D}^{-\frac12}u$ is an eigenvector of
    $A_{rm}$ with the same eigenvalue, i.e., $A_{rm}\tilde{D}^{-\frac12}u = \lambda \tilde{D}^{-\frac12} u$.
    \item The eigenvector associated with the largest eigenvalue $\lambda_1=1$ of $A_{sys}$ is $u_1=\tilde{D}^{\frac12}$, while 
    for $A_{rw}$, it is $A_{rw}\mathbf{1} = \lambda_1\mathbf{1} =\mathbf{1}$.
\end{itemize}

\paragraph{Mean-subtraction for $A_{sys}$.}Let first discuss the graph convolution operator $A_{sys}$,
\begin{align}
A_{sys} = \tilde{D}^{-\frac12}(I+A)\tilde{D}^{-\frac12}=U\Lambda U^\top = \sum_{i=1}^su_i \lambda_i u_i^\top
\end{align}
and the signal $x$ as 
\begin{equation}
    x =\epsilon + \sum_{i=1}^sc_i\cdot u_i
\end{equation}
Then we apply the Laplacian smoothing $k$ times,
\begin{align}
    A_{sys}^k x &= \left(\sum_{i=1}^su_i \lambda_i u_i^\top\right)^k \left(\epsilon + \sum_{i=1}^sc_i\cdot u_i\right) \notag\\
    & = \left(\sum_{i=1}^su_i \lambda_i^k u_i^\top\right) \left(\epsilon + \sum_{i=1}^sc_i\cdot u_i\right) = \sum_{i=1}^sc_i\lambda_i^k u_i.
\end{align}
which tells that $\displaystyle{\lim_{k \to \infty}} A_{sym}^kx \propto u_1 = \tilde{D}^{\frac12}\textbf{1}$. The mean-subtraction trick on $A_{sym}$ is of
a factor $\tilde{D}^{\frac12}$ (suppose mapping $f(x) = \tilde{D}^{\frac12}x$ and inverse mapping $f^{-1}(x) = \tilde{D}^{-\frac12}x$),
\begin{equation}
    x_{new} \leftarrow f^{-1}(f(x)-f(x_{mean})) = \tilde{D}^{-\frac12}(1- \frac{\textbf{1} \textbf{1}^\top}{n})\tilde{D}^{\frac12}x = x - \frac{\tilde{D}^{\frac12}\textbf{1} \textbf{1}^\top \tilde{D}^{-\frac12}x}{n}.
\end{equation}
Therefore, after one layer of mean-subtraction, the signal $x$ would be,
\begin{align}
    x_{new}&\leftarrow f^{-1}(f(x)-f(x_{mean})) \notag\\
    &= \tilde{D}^{-\frac12}(1- \frac{\textbf{1} \textbf{1}^\top}{n})\tilde{D}^{\frac12}\left(\epsilon + \sum_{i=1}^sc_i\cdot u_i\right) \notag\\
    &= \tilde{D}^{-\frac12}(1- \frac{\textbf{1} \textbf{1}^\top}{n})\tilde{D}^{\frac12}\sum_{i=1}^sc_i\cdot u_i \notag\\
    &= \sum_{i=1}^s \tilde{D}^{-\frac12}(1- \frac{\textbf{1} \textbf{1}^\top}{n})\tilde{D}^{\frac12} c_i\cdot u_i \notag\\
    &= \sum_{i=1}^s c_i\cdot u_i -\sum_{i=1}^s \tilde{D}^{-\frac12}\frac{\textbf{1} \textbf{1}^\top}{n}\tilde{D}^{\frac12} c_i\cdot u_i \notag\\
    &= \sum_{i=1}^s c_i\cdot u_i -\sum_{i=1}^1 c_i\cdot u_i = \sum_{i=2}^s c_i\cdot u_i
\end{align}
which eliminate the dominant effect of $u_1$.

\paragraph{Mean-subtraction for $A_{rw}$.}Then for the graph convolution operator $A_{rw}$, we could do the similar decomposition,

\begin{align}
A_{rw} &= \tilde{D}^{-1}(I+A) \notag\\
&= \tilde{D}^{-\frac12}\left(\tilde{D}^{-\frac12}(I+A)\tilde{D}^{-\frac12}\right)\tilde{D}^{\frac12} \notag\\
&=\tilde{D}^{-\frac12}\left(U\Lambda U^\top\right)\tilde{D}^{\frac12} \notag\\
&= \tilde{D}^{-\frac12}\left(\sum_{i=1}^su_i \lambda_i u_i^\top\right)\tilde{D}^{\frac12}
\end{align}
and for the signal $x$ into $\{\tilde{D}^{-\frac12}u\}$ space as 
\begin{equation}
    x =\epsilon + \sum_{i=1}^sc_i\cdot \tilde{D}^{-\frac12} u_i
\end{equation}
Similar we apply the Laplacian smoothing $k$ times,
\begin{align}
    A_{rw}^k x &= \left(\tilde{D}^{-\frac12}\left(\sum_{i=1}^su_i \lambda_i u_i^\top\right)\tilde{D}^{\frac12}\right)^k \left(\epsilon + \sum_{i=1}^sc_i\cdot \tilde{D}^{-\frac12}u_i\right) \notag\\
    & = \underbrace{\left(\tilde{D}^{-\frac12}\left(\sum_{i=1}^su_i \lambda_i u_i^\top\right)\tilde{D}^{\frac12}\right)\cdots \left(\tilde{D}^{-\frac12}\left(\sum_{i=1}^su_i \lambda_i u_i^\top\right)\tilde{D}^{\frac12}\right)}_{k ~terms} \left(\epsilon + \sum_{i=1}^sc_i\cdot \tilde{D}^{-\frac12}u_i\right) \notag\\
    & = \tilde{D}^{-\frac12}\left(\sum_{i=1}^su_i \lambda_i^k u_i^\top\right)\tilde{D}^{\frac12} \left(\epsilon + \sum_{i=1}^sc_i\cdot \tilde{D}^{-\frac12}u_i\right)= \sum_{i=1}^sc_i\lambda_i^k \tilde{D}^{-\frac12}u_i.
\end{align}
which tells that $\displaystyle{\lim_{k \to \infty}} A_{rw}^kx \propto \tilde{D}^{-\frac12}u_1 = \textbf{1}$. The mean-subtraction trick on $A_{rw}$ is
\begin{equation}
    x_{new} \leftarrow x-x_{mean} = x - \frac{\textbf{1} \textbf{1}^\top x}{n} = x - \langle x, \bar{u_1}\rangle\cdot \bar{u_1}
\end{equation}
Therefore, after one layer of mean-subtraction, the signal $x$ would be,
\begin{align}
    x_{new}&\leftarrow x-x_{mean} \notag\\
    &= (1- \frac{\textbf{1} \textbf{1}^\top}{n})\left(\epsilon + \sum_{i=1}^sc_i\cdot \tilde{D}^{-\frac12}u_i\right) \notag\\
    &= (1- \frac{\textbf{1} \textbf{1}^\top}{n})\sum_{i=1}^sc_i\cdot \tilde{D}^{-\frac12}u_i \notag\\
    &= \sum_{i=1}^s (1- \frac{\textbf{1} \textbf{1}^\top}{n})c_i\cdot \tilde{D}^{-\frac12}u_i \notag\\
    &= \sum_{i=1}^s c_i\cdot \tilde{D}^{-\frac12}u_i -\sum_{i=1}^s \frac{\textbf{1} \textbf{1}^\top}{n}c_i\cdot \tilde{D}^{-\frac12}u_i \notag\\
    &= \sum_{i=1}^s c_i\cdot \tilde{D}^{-\frac12}u_i -\sum_{i=1}^1 c_i\cdot \tilde{D}^{-\frac12}u_i = \sum_{i=2}^s c_i\cdot \tilde{D}^{-\frac12}u_i
\end{align}
which eliminate the dominant effect of $\tilde{D}^{-\frac12}u_1=\mathbf{1}$.

\paragraph{Remark 4.}  So far, we discuss the one layer mean-subtraction for both $A_{sys}$ and $A_{rw}$ and also the powering effect of $A_{sys}$ and $A_{rw}$ on arbitrary signal $x$ ($c_1$ is non-zero). Although we have show that one layer of mean-subtraction could eliminate the dominant eigenvector (once and for all). However, in the main text, we discuss that
in the non-linear deep GCN architecture, which means after the ReLU activation function, the effect of dominant eigenvector may still appear. Therefore, we need mean-subtraction layer after applying activation function and iteratively eliminate $u_1$ or $\tilde{D}^{-\frac12}u_1$. Due to the powering effect, they will finally approximate the Fiedler vector,
\begin{equation}
    \displaystyle{\lim_{k \to \infty}} {[A_{rw}^kx]}_{mean-subtraction} \propto \tilde{D}^{-\frac12}u_2 ~~~~~\mbox{and}~~~~~ \displaystyle{\lim_{k \to \infty}} {[A_{sys}^kx]}_{mean-subtraction} \propto u_2
\end{equation}

\section{\textit{Karate} Demonstration} \label{sec:AppendixF}

\paragraph{Mean-subtraction for \textit{Karate}.}
We use the mean-subtraction trick on \textit{Karate} data. The experiment setting is as follows: we randomly assign 2-dimensional feature vector for each node and apply Laplacian smoothing $k$ times ($k=0,5,20,100$) with normalized random-walk adjacency operator $A_{rw}$. For each $k$, we visualize the feature vector of each node after scaling the dimension by the largest absolute value in that dimension (i.e., f = f / max(abs(f))). From ground truth, each color indicates a class and we manually add them to help with the visualization. 

It is impressive that with mean-subtraction, nodes
are almost well-separated during multi-layer Laplacian smoothing. As is stated in the main text, the reason is that mean-subtraction magnifies the Fiedler vector and achieves a pre-separation effect.
\begin{figure}[htbp]
	\centering
	\includegraphics[width=5.5in]{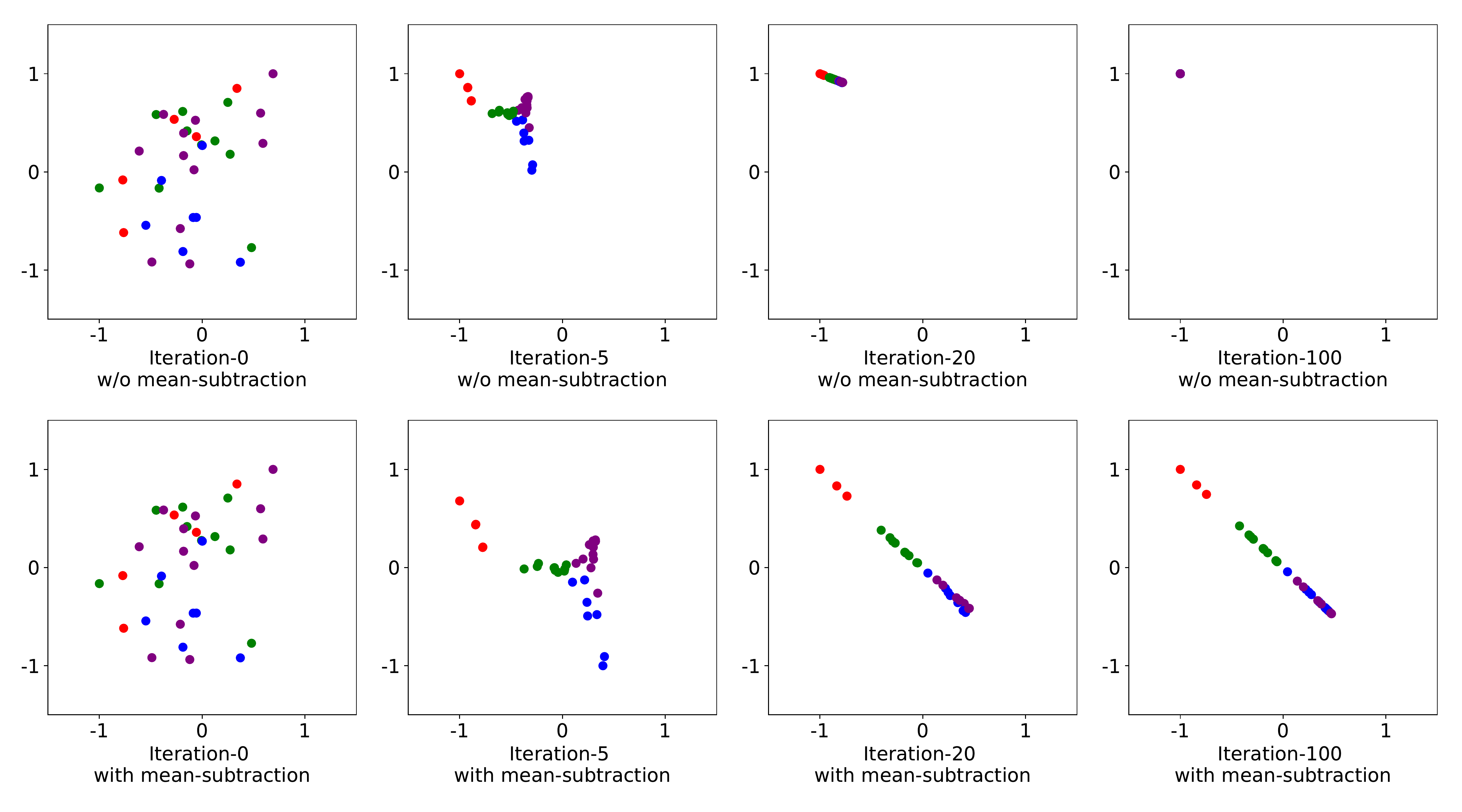}
	\caption{Laplacian Smoothing for \textit{Karate} with or w/o Mean-subtraction}
	\label{fig:karate2}
\end{figure}

\paragraph{The Cosine Similarity.} Suppose the feature matrix after the $l$-th layer is $X^{(l)} \in \mathbb{R}^{n\times m}$,
\begin{equation}
   X^{(l)} = \begin{bmatrix}
   x_{11} & \cdots & x_{1m} \\
   \vdots & \ddots & \\
   x_{n1} &     &  x_{nm}\end{bmatrix}
\end{equation}
we use $X_i,~i=1..n$ to denote the rows of $X^{(l)}$ and use $X^j,~j=1..m$ to denote the cols of $X^{(l)}$. $\{X_i\}$ is row vector and $\{X^j\}$ is column vector. The feature-wise smoothing (cosine similarity) is given by the averaging absolute value of the following matrix,
\begin{equation}
    \mbox{avg} \begin{bmatrix}
    \frac{X^{1\top} X^1}{\|X^1\|\|X^1\|} & \frac{X^{1\top} X^2}{\|X^1\|\|X^2\|} & \cdots & \frac{X^{1\top} X^m}{\|X^1\|\|X^m\|} \\
    \frac{X^{2\top} X^1}{\|X^2\|\|X^1\|} & \frac{X^{2\top} X^2}{\|X^2\|\|X^2\|} & \cdots & \frac{X^{2\top} X^m}{\|X^2\|\|X^m\|} \\
    \vdots & \vdots & \ddots & \\
    \frac{X^{m\top} X^1}{\|X^m\|\|X^1\|} & \frac{X^{m\top} X^2}{\|X^m\|\|X^2\|} &  & \frac{X^{m\top} X^m}{\|X^m\|\|X^m\|}
    \end{bmatrix}
\end{equation}
The maximum possible score is $1$ if all the entry are either $1$ or $-1$, which means all of they are entirely on the same direction. 

Similarly, the node-wise smoothing (cosine similarity) is given by the averaging absolute value of the following matrix,
\begin{equation}
    \mbox{avg} \begin{bmatrix}
    \frac{X_1 X_1^\top}{\|X_1\|\|X_1\|} & \frac{X_1 X_2^\top}{\|X_1\|\|X_2\|} & \cdots & \frac{X_1X_n^\top}{\|X_1\|\|X_n\|} \\
    \frac{X_2 X_1^\top}{\|X_2\|\|X_1\|} & \frac{X_2 X_2^\top}{\|X_2\|\|X_2\|} & \cdots & \frac{X_2 X_n^\top}{\|X_2\|\|X_n\|} \\
    \vdots & \vdots & \ddots & \\
    \frac{X_n X_1^\top}{\|X_n\|\|X_1\|} & \frac{X_n X_2^\top}{\|X_n\|\|X_2\|} &  & \frac{X_n X_n^\top}{\|X_n\|\|X_n\|}
    \end{bmatrix}
\end{equation}
The maximum possible score is $1$ if all the entry are either $1$ or $-1$, which means all of they are entirely on the same direction.

\section{Experimental Details and More} \label{sec:additionalexp}

\subsection{Additional Experiments}
\begin{figure}[htbp!]
	\centering
	\includegraphics[width=5.5in]{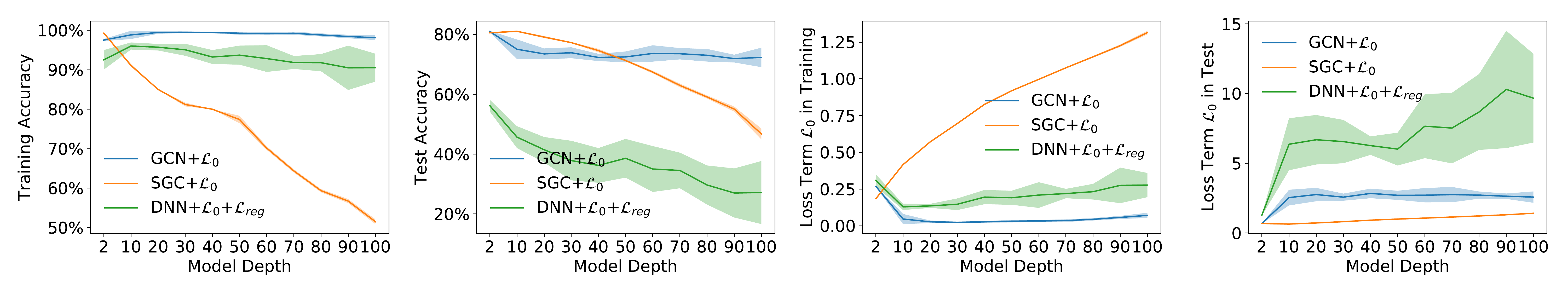}
	\caption{Comparison of Deep GCN, Deep SGC and DNN on \textit{Cora}}
	\label{fig:GCNvsSGC}
\end{figure}

\begin{figure}[htbp!]
	\centering
	\includegraphics[width=5.5in]{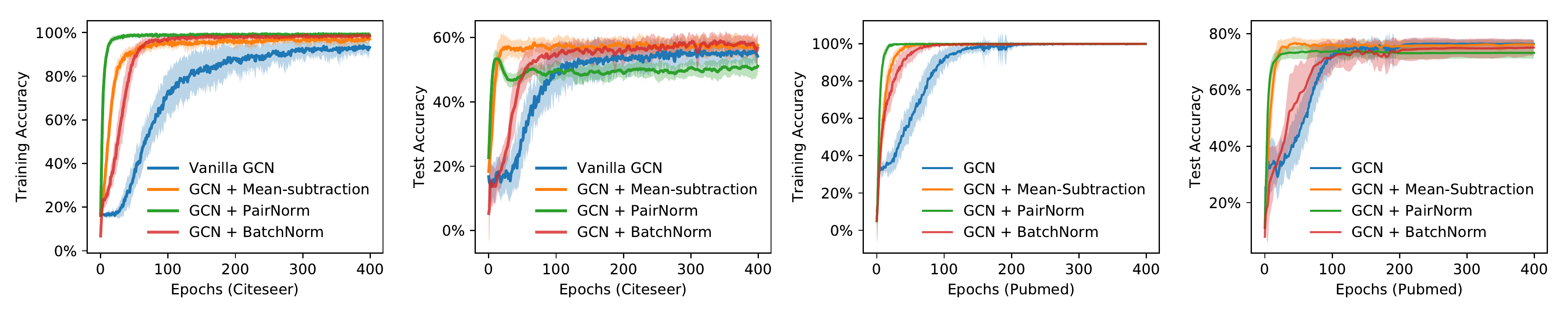}
	\caption{Comparison of Different Tricks in Training Deep GCNs on \textit{Citeseer, Pubmed}.}
	\label{fig:mean-sub}
\end{figure}

\begin{figure}[thbp!]
	\centering
	\includegraphics[width=5.5in]{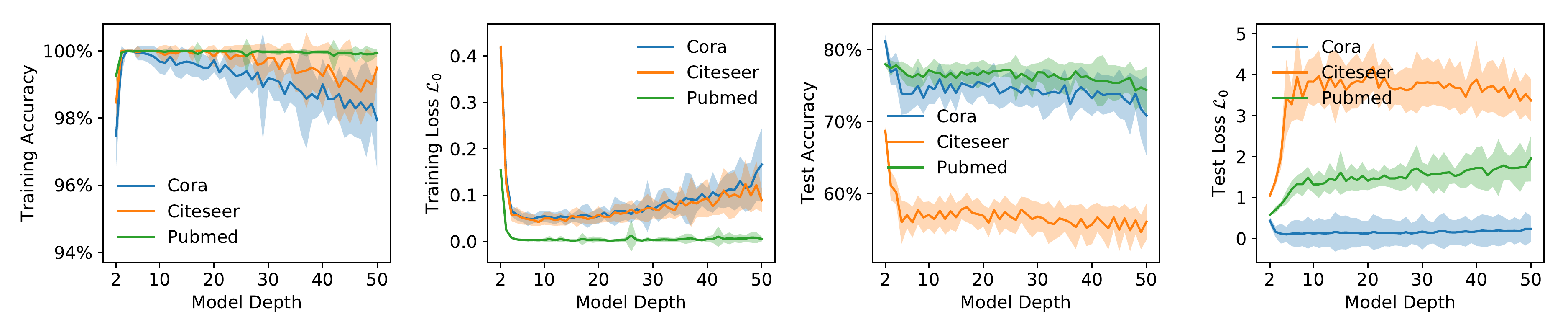}
	\caption{Training Deep GCNs for $2\sim50$ layers on \textit{Cora, Citeseer, Pubmed}.}
	\label{fig:threeset}
\end{figure}

\begin{figure}[t]
	\centering
	\includegraphics[width=5.5in]{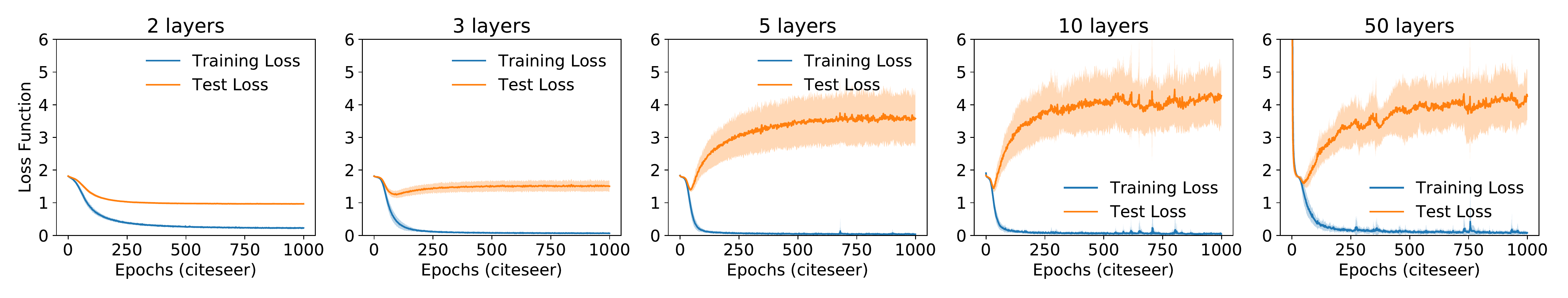}
	\caption{Training and Test Curve with 2-, 3-, 5-, 10-, 50-layer GCNs on \textit{Citeseer}.}
	\label{fig:loss-citeseer}
\end{figure}

\begin{figure}[t]
	\centering
	\includegraphics[width=5.5in]{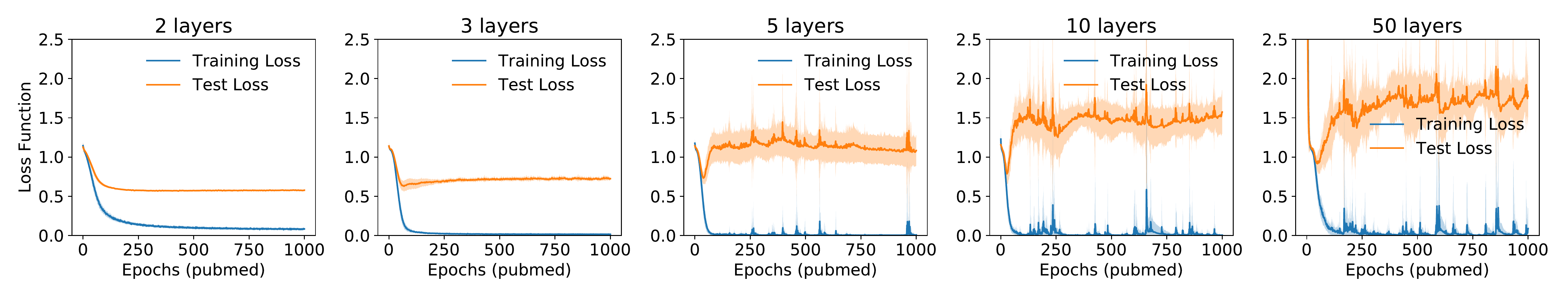}
	\caption{Training and Test Curve with 2-, 3-, 5-, 10-, 50-layer GCNs on \textit{Pubmed}.}
	\label{fig:loss-pubmed}
\end{figure}

Note that all the experiments are conducted for 20 times.
\begin{itemize}
	\item We compare deep GCNs (with $\mathcal{L}_0$), deep SGC (with $\mathcal{L}_0$) and DNN (with $\mathcal{L}_0 + \gamma \mathcal{L}_{reg}$) on \textit{Cora} with different depths  in Figure~\ref{fig:GCNvsSGC};
	\item Additional mean-subtraction evaluations for \textit{Citeseer, Pubmed} in  Figure~\ref{fig:mean-sub};
	\item Training Deep GCNs for $2\sim50$ layers on \textit{Cora, Citeseer, Pubmed}. We show the training/test loss and accuracy curves in Figure~\ref{fig:threeset};
	\item We compute the training and test loss of the vanilla GCN models of 2-, 3-, 5-, 10-, 50-layer with 1000 epochs on \textit{Citeseer} is reported in Figure~\ref{fig:loss-citeseer};
	\item We compute the training and test loss of the vanilla GCN models of 2-, 3-, 5-, 10-, 50-layer with 1000 epochs on \textit{Pubmed} is reported in Figure~\ref{fig:loss-pubmed}.
\end{itemize} 

%\begin{figure}[htbp!]
%	\centering
%	\includegraphics[width=4.5in]{fig/depthmodel.pdf}
%	\caption{Loss and Accuracy of Training and Test over \textit{Cora}}
%	\label{fig:modeldepth}
%\end{figure}
%
%\begin{figure}[htbp!]
%	\centering
%	\includegraphics[width=4.5in]{fig/depthmodel2.pdf}
%	\caption{Loss and Accuracy of Training and Test over \textit{Citeseer}}
%	\label{fig:modeldepth2}
%\end{figure}
%
%\begin{figure}[htbp!]
%	\centering
%	\includegraphics[width=4.5in]{fig/depthmodel3.pdf}
%	\caption{Loss and Accuracy of Training and Test over \textit{PubMed}}
%	\label{fig:modeldepth3}
%\end{figure}

\end{document}